\documentclass{article}

\usepackage{microtype}
\usepackage{graphicx}
\usepackage{booktabs} 

\usepackage{hyperref}

\usepackage[accepted]{icml2025}

\usepackage{amsmath}
\usepackage{amssymb}
\usepackage{mathtools}
\usepackage{amsthm}

\usepackage[capitalize,noabbrev]{cleveref}

\theoremstyle{plain}

\theoremstyle{definition}

\theoremstyle{remark}

\usepackage[textsize=tiny]{todonotes}

\usepackage{times}
\usepackage{latexsym}
\usepackage{booktabs}
\usepackage{adjustbox}
\usepackage{makecell}
\usepackage{multicol}
\usepackage{multirow}
\usepackage{hhline}
\usepackage{tikz}
\usepackage{xcolor}
\usepackage{colortbl}
\usepackage{natbib}
\usepackage{subcaption}
\usepackage{pifont}
\usepackage{url}
\usepackage{amsmath}
\usepackage{relsize}
\usepackage{caption}
\usepackage[normalem]{ulem}
\useunder{\uline}{\ul}{}
\usepackage{hyperref}
\usepackage{cleveref}
\usepackage{enumitem}
\usepackage{wrapfig}
\usepackage{subcaption}

\usepackage[T1]{fontenc}

\usepackage[utf8]{inputenc}

\usepackage{microtype}

\usepackage{inconsolata}

\usepackage{graphicx}

\definecolor{myGreen}{RGB}{127,210,85}
\definecolor{myOrange}{RGB}{242,154,66}
\definecolor{myYellow}{RGB}{247,223,65}
\definecolor{myRed}{RGB}{232,80,43}
\definecolor{myViolet}{RGB}{162,57,102}
\definecolor{myBlue}{HTML}{4686f3}
\definecolor{myYellowv2}{HTML}{E6C802}
\definecolor{myOrangev2}{HTML}{ED8E55}
\definecolor{MyGreenv2}{HTML}{009B55}
\definecolor{MyRedv2}{HTML}{c22f2f}
\definecolor{LinkPink}{HTML}{df1a7d}

\definecolor{CustomColor1}{RGB}{245,210,209}
\definecolor{CustomColor2}{RGB}{230,232,245} 
\definecolor{CustomColor3}{RGB}{235,235,237} 
\definecolor{ForestGreen}{HTML}{009B55}
\definecolor{OrangeRed}{HTML}{c22f2f}

\newcommand{\dataname}{\textsc{AAAR-1.0}}
\newcommand{\tasknameequationlong}{\textsc{EquationInference}}
\newcommand{\tasknamereviewlong}{\textsc{PaperWeakness}}
\newcommand{\tasknameexperimentlong}{\textsc{ExperimentDesign}}
\newcommand{\tasknamemetareviewlong}{\textsc{ReviewCritique}}

\newcommand{\tasknameequation}{\textsc{EqInfer}}
\newcommand{\tasknamereview}{\textsc{Weakness}}
\newcommand{\tasknameexperiment}{\textsc{ExpDesign}}
\newcommand{\tasknamemetareview}{\textsc{ReviewCritique}}
\newcommand{\MetricReviewIDF}{ITF-IDF}
\newcommand{\MetricReviewRecall}{S-Recall}
\newcommand{\MetricReviewPrec}{S-Precision}
\newcommand{\MetricReviewF}{S-F$_{1}$}
\newcommand{\MetricExpMatch}{S-Match}
\newcommand{\MetricExpRecall}{En-Recall}
\newcommand{\MetricExpPrec}{En-Precision}
\newcommand{\MetricExpF}{En-F$_{1}$}
\newcommand{\AISci}{\textsc{AI-SCI}}

\definecolor{ShallowBlue}{HTML}{ddf2ff}
\definecolor{ShallowPink}{HTML}{fadfdc}
\definecolor{ShallowGreen}{HTML}{cff0da}
\definecolor{ShallowOrange}{HTML}{fff1b9}
\newcommand{\ColoredEQ}{\colorbox{ShallowGreen}{\raisebox{0pt}[0.7\height][0.4\depth]\tasknameequation}}
\newcommand{\ColoredEXP}{\colorbox{ShallowPink}{\raisebox{0pt}[0.7\height][0.4\depth]\tasknameexperiment}}
\newcommand{\ColoredReview}{\colorbox{ShallowBlue}{\raisebox{0pt}[0.7\height][0.4\depth]\tasknamereview}}
\newcommand{\ColoredMetaReview}{\colorbox{ShallowOrange}{\raisebox{0pt}[0.7\height][0.4\depth]\tasknamemetareview}}

\newcommand{\MyUpArrow}[1]{%
  \textcolor{ForestGreen}{$\uparrow$ #1}%
}
\newcommand{\MydownArrow}[1]{%
  \textcolor{OrangeRed}{$\downarrow$ #1}%
}

\defcitealias{gemma_2024}{(Gemma Team, 2024)}
\defcitealias{qwen2.5}{(Qwen Team, 2024)}
\defcitealias{team2023gemini}{(Gemini Team, 2024)}

\crefname{section}{§}{§§}
\definecolor{darkblue}{rgb}{0, 0, 0.5}
\hypersetup{colorlinks=true,citecolor=darkblue, linkcolor=darkblue, urlcolor=darkblue}

\icmltitlerunning{\dataname: Assessing  AI's Potential to Assist Research}

\begin{document}

\twocolumn[
\icmltitle{\includegraphics[width=0.3in]{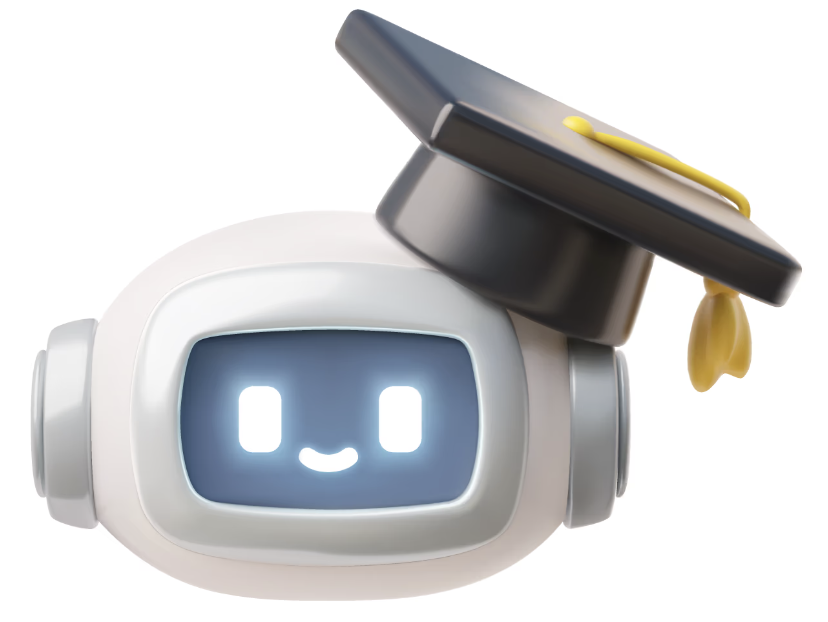}~\dataname: Assessing  AI's Potential to Assist Research}



\icmlsetsymbol{equal}{*}

\begin{icmlauthorlist}
\icmlauthor{Renze Lou}{psu} 
\icmlauthor{Hanzi Xu}{netflix}
\icmlauthor{Sijia Wang}{ucd}
\icmlauthor{Jiangshu Du}{uic}
\icmlauthor{Ryo Kamoi}{psu}
\icmlauthor{Xiaoxin Lu}{psu}
\icmlauthor{Jian Xie}{meta}
\icmlauthor{Yuxuan Sun}{meta}
\icmlauthor{Yusen Zhang}{psu}
\icmlauthor{Jihyun Janice Ahn}{psu}
\icmlauthor{Hongchao Fang}{psu}
\icmlauthor{Zhuoyang Zou}{psu}
\icmlauthor{Wenchao Ma}{psu}
\icmlauthor{Xi Li}{uab}
\icmlauthor{Kai Zhang}{osu}
\icmlauthor{Congying Xia}{meta}
\icmlauthor{Lifu Huang}{ucd}
\icmlauthor{Wenpeng Yin}{psu}
  
\end{icmlauthorlist}

\icmlaffiliation{psu}{Pennsylvania State University;}
\icmlaffiliation{netflix}{Netflix;}
\icmlaffiliation{ucd}{University of California, Davis;}
\icmlaffiliation{uic}{University of Illinois Chicago;}
\icmlaffiliation{uab}{University of Alabama at Birmingham;}
\icmlaffiliation{osu}{Ohio State University}
\icmlaffiliation{meta}{Individual Researcher;}

\icmlcorrespondingauthor{Renze Lou}{renze.lou@psu.edu}
\icmlcorrespondingauthor{Wenpeng Yin}{wenpeng@psu.edu}

\icmlkeywords{Machine Learning, ICML}

\vskip 0.3in
]



\printAffiliationsAndNotice{Work done prior to Jiangshu joining Amazon.}  

\begin{abstract}
Numerous studies have assessed the proficiency of AI systems, particularly large language models (LLMs), in facilitating everyday tasks such as email writing, question answering, and creative content generation.
However, researchers face unique challenges and opportunities in leveraging LLMs for their own work, such as brainstorming research ideas, designing experiments, and writing or reviewing papers. 
In this study, we introduce \dataname, a benchmark dataset designed to evaluate LLM performance in four fundamental, expertise-intensive research tasks: (i) \tasknameequationlong, assessing the correctness of equations based on the contextual information in paper submissions; (ii) \tasknameexperimentlong, designing experiments to validate research ideas and solutions; (iii) \tasknamereviewlong, identifying weaknesses in paper submissions; and (iv) \tasknamemetareviewlong, identifying each segment in human reviews is deficient or not. \dataname~differs from prior benchmarks in two key ways: first, it is explicitly research-oriented, with tasks requiring deep domain expertise; second, it is researcher-oriented, mirroring the primary activities that researchers engage in on a daily basis. An evaluation of both open-source and closed-source LLMs reveals their potential as well as limitations in conducting sophisticated research tasks. 
We will keep iterating \dataname~to new versions. 
\textcolor{LinkPink}{Project Webpage:} ~\textbf{
\hypersetup{urlcolor=LinkPink}\url{https://renzelou.github.io/AAAR-1.0/}}

\end{abstract}

\begin{figure*}[t]
 \setlength{\abovecaptionskip}{2pt}
 \centering
    \includegraphics[width=0.94\linewidth]{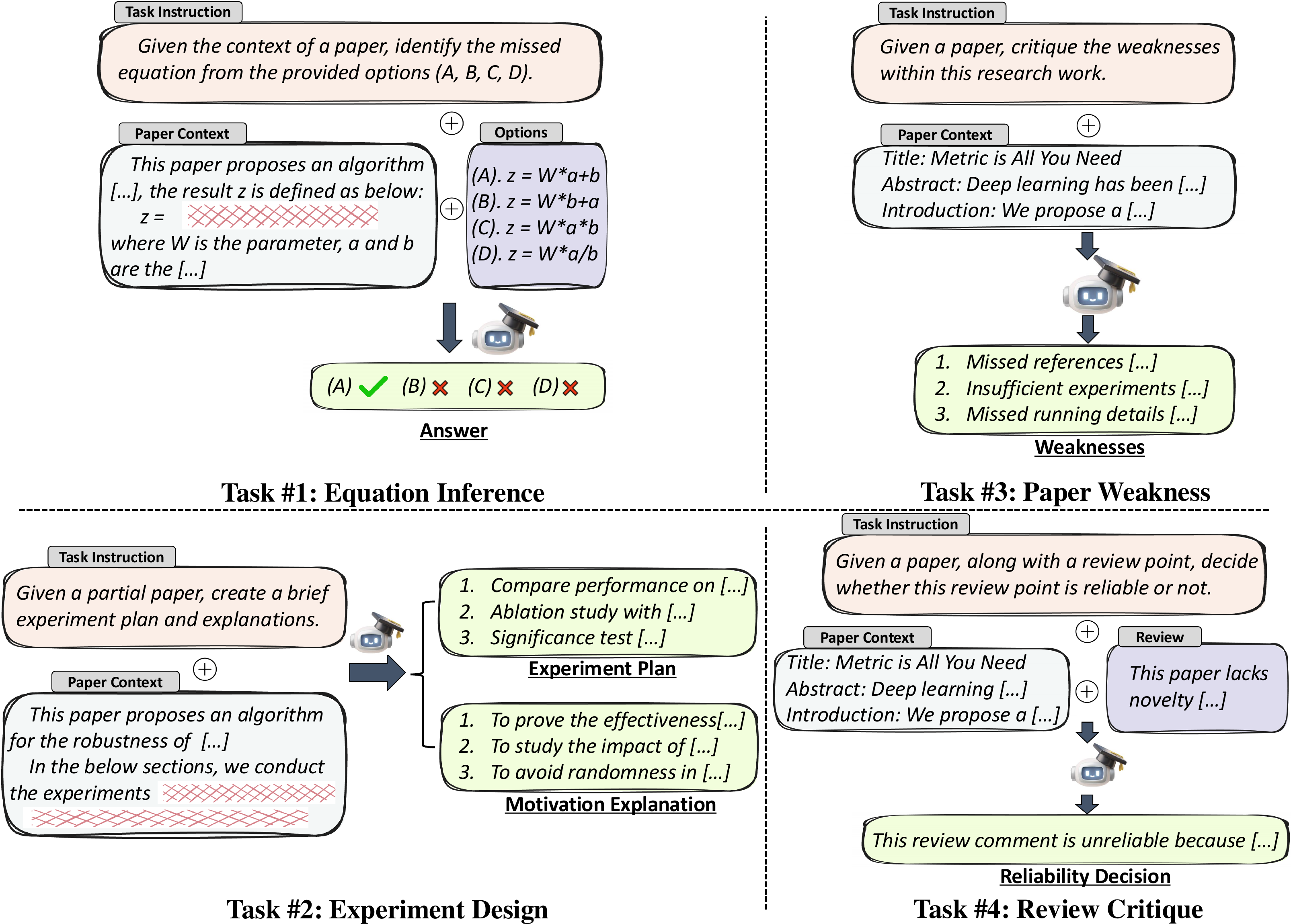}
    \caption{The input-output illustration of four tasks in the proposed~\dataname~benchmark.}
    \label{fig:intro_three_tasks}
\end{figure*}

\section{Introduction}


Although AI has brought transformative changes to various aspects of life, its impact on researchers unfolds in a nuanced manner. On the one hand, AI assists in various research disciplines, such as Social Science \citep{DBLPNeumanCY23}, Finance \citep{ai4finance}, Medicine \citep{rakhimov2022artificial}, GeoScience \citep{praskievicz2018river}, etc., significantly expediting academic processes. However, many of these applications are superficial, often limited to data-driven clustering or classification. On the flip side, the AI era poses challenges for researchers. Despite its ability to streamline some activities, researchers still face demanding, cognitively intensive tasks such as staying current through extensive paper reading, rapidly generating ideas in response to fast-paced advancements, conducting rigorous experiments to substantiate claims, and managing an increasing volume of peer reviews. Then a question looms: \emph{How effectively can AI assist researchers in tasks that are domain-specific, expertise-demanding, and reasoning-intensive?}

Existing works proved the promising potential for using LLMs in assisting AI research. \citet{si2024can} conducted a large-scale human study and found that LLMs can generate creative research ideas. \citet{lu2024aiscientist} proposed an autonomous agent to handle complicated research workflow and write a whole research paper.
However, most of these works focus on addressing highly subjective problems that require a high degree of expertise, making evaluation laborious and hard to reproduce. This underscores the need for a comprehensive benchmark that rigorously assesses LLMs' capabilities in expertise-intensive research activities.

To this end, in this work, we introduce \dataname, a novel benchmark that aims to comprehensively assess the LLMs' capacity on expert-level research tasks. As illustrated in Figure~\ref{fig:intro_three_tasks}, \dataname~decomposes four distinct expert-level AI research tasks from the researcher's daily activities, including i) \tasknameequationlong, investigating whether the LLMs can infer the equation correctness based on the paper context; ii) \tasknameexperimentlong, validating LLMs' ability on designing reliable experiments for a research idea; iii) \tasknamereviewlong, testing the quality of weaknesses discovered by LLMs from paper drafts; and iv) \tasknamemetareviewlong, investigating whether LLMs can identify and explain the deficient/unreliable human-written paper reviews. To ensure data quality, senior AI researchers with extensive domain expertise perform data annotation for \dataname, followed by rigorous multi-round data examination and filtering.
%
All tasks require models to possess strong domain knowledge covering various cutting-edge research findings, as well as expert-level research experience, to the extent that even humans need substantial research accumulation to tackle the tasks we designed.
Crucially, tasks here are singular, stand-alone challenges (with clear input and output expectations) rather than a complicated task chain~\citep{li2024mlr,lu2024aiscientist}, providing a more transparent assessment of the model's intermediate output. 

Benefiting from the proposed automatic metrics, we conduct extensive experiments across numerous mainstream LLMs, where we find that:

\begin{itemize}[leftmargin=2em]
\itemsep0em 
    \item With a random guess baseline of 40\% F$_1$, the performance of most LLMs on \tasknameequation~hovers just slightly above chance, with the top models reaching around 46\%. This highlights the difficulty of the task, despite its reliance primarily on local context reasoning.
    \item In \tasknameexperiment, LLM-designed experiments are innovative and more diverse than those by humans; however, many are trivial, lack feasibility, and stray from the original research objectives.
    \item In \tasknamereviewlong, LLM-identified weaknesses often lack depth and specificity, making them broadly applicable and less useful for providing feedback on paper drafts.
    \item In \tasknamemetareviewlong, LLMs struggle to effectively identify deficient human reviews, indicating limited usefulness in assisting meta-reviewers in evaluating the quality of individual paper reviews.
    LLMs generated weakness lacks sufficient domain-specific knowledge, model always tends to generate some vague weaknesses that can be applied to any other papers.
\end{itemize}


\section{Related Work}
\label{appendix:related_work}


\paragraph{LLMs for AI Research.}
With the rapid evolution of pertaining techniques, LLMs are found to be useful in assisting various research disciplines~\citep{yu2024llasmol,labrak2024biomistral}, particularly in AI research, such as generating novel research ideas~\citep{kumar2024can,yu2024researchtown}, reviewing research draft~\citep{gao2024reviewer2,du2024llms,liang2024can,zhu2025deepreview}, and writing scientific papers~\citep{chamoun2024automated,lu2024aiscientist,weng2024cycleresearcher}. For example, \citet{si2024can} conducted a large-scale human investigation on LLM-generated research ideas and found that LLMs can generate novel ideas compared with humans while lacking feasibility. \citet{du2024llms} found that while LLMs are effective at summarizing papers, they tend to overly trust the authors' claimed strengths and struggle to identify weaknesses specific to the paper. Furthermore, some works try to employ LLMs to solve more complicated research tasks that are composed of multiple steps~\citep{li2024mlr,li2023traineragent,tang2023ml}. Notably, \citet{lu2024aiscientist} proposed \textsc{AI-Scientist}, an autonomous agent framework that can handle a series of challenging research tasks consecutively, including generating research ideas, coming up with the corresponding experiments along with the implementations, and then writing the final research paper --- exactly how human conduct a whole research pipeline. However, there is still a lack of systematic evaluations and quantitative analyses on the LLMs' (intermediate) output of each single-step research task. Accordingly, our work focuses on building a benchmark consisting of individual research steps with clear input-output expectations, making it suitable for comprehensive LLM evaluation. Moreover, \textbf{we emphasize that relying on LLMs to fully replace human effort might compromise academic integrity}. While our benchmark primarily serves an educational purpose --- LLMs assist junior researchers by providing imperfect but insightful ideas, rather than by governing the entire research process.

\paragraph{Benchmarks for AI Research Tasks.}
Existing ``LLM assists research'' benchmarks mainly focus on the implementation and execution part of the research pipeline~\citep{lu2024aiscientist,chen2024scienceagentbench,li2024mlr,chan2024mle}. For instance, \citet{huang2024mlagentbench} proposed MLAgentBench to test the LLMs' capacity for writing project code and training the ML models, where the evaluation metric is the test performance of the models trained by LLMs.
However, real-world AI research activities are diverse and some of them are hard to assess for quality, such as generating research ideas, which requires intensive manual assessment~\citep{si2024can,liang2024can}.
Our work centers on tasks that emphasize a comprehensive mastery of the scientific research field and core elements of a researcher’s daily workload, and we try to build curated task-specific metrics for every single task for a more efficient and accurate LLMs appraisal.

\begin{figure*}[t]
 \setlength{\belowcaptionskip}{-13pt}
 \setlength{\abovecaptionskip}{-1pt}
	\begin{center}
		\centering
        \includegraphics[width=0.96\linewidth]{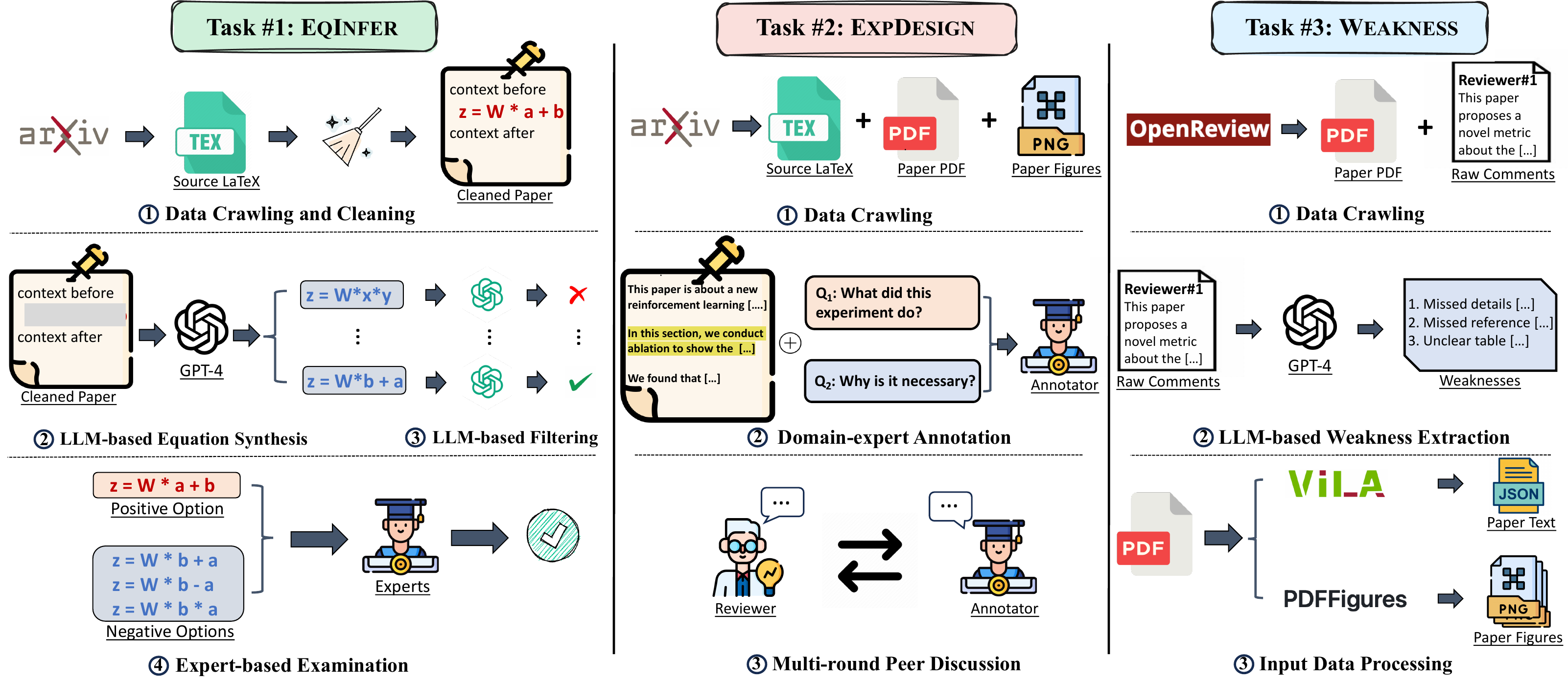}
	\end{center}
	\caption{Data construction workflows of the three tasks in \includegraphics[width=0.21in]{figures/robot_expert.png}~\dataname.}
\label{fig:data_collection}
\end{figure*}

\section{\dataname}
\label{sec:data_collection}

%


Figure~\ref{fig:data_collection} provides a data construction overview. In the following sections, we elaborate on the data collection details, including \cref{subsec:data_collection_eq}~{\tasknameequationlong}~(\textbf{\ColoredEQ}), \cref{subsec:data_collection_exp}~{\tasknameexperimentlong}~(\textbf{\ColoredEXP}), and \cref{subsec:data_collection_review}~{\tasknamereviewlong}~(\textbf{\ColoredReview}),
and \cref{subsec:data_collection_metareview}~{\textbf{\ColoredMetaReview}}.



\subsection{\tasknameequationlong}
\label{subsec:data_collection_eq}


Crafting a correct scientific equation in paper writing or validating an equation in paper reviewing is challenging, as it requires a thorough understanding of an algorithm or the intricate relationships among numerous variables. Directly prompting LLMs to generate equations proves overly demanding. Therefore,  this work formulates \textbf{\ColoredEQ} (Figure~\ref{fig:intro_three_tasks}) as a binary inference task.\footnote{\tasknameequation~also facilitates a multiple-choice QA setting; while we find a binary inference is more challenging for LLMs.}

\paragraph{\ding{172} Data crawling and cleaning.}

For the data source, we adopt the pre-compilation LaTeX code for two reasons: i) existing PDF parsing tools, such as PyMuPDF and PaperMage~\citep{lo2023papermage}, can introduce considerable noise to the parsed equation text; ii) considering most of exiting LLMs are capable with processing LaTeX code, using LaTeX source instead of parsed text can be more accurate and provide LLMs with richer information.
Meanwhile, we only crawl those peer-reviewed papers accepted by top-tier conferences to avoid using low-quality human-written equations. Accordingly, we first obtain the accepted paper list from ACL Anthology, from year \texttt{2019} to \texttt{2023}. Next, we search each paper on arXiv to crawl its LaTeX source (if it exists). Finally, we get a total of 1,762 papers' source LaTeX packages.
We then clean the LaTeX sources by deleting all the comments and combining multiple cross-referred \texttt{.tex} files into a main file. Afterward, we use regex to randomly extract (at most) 3 equations' code snippets per paper, resulting in 3,877 human-written equations.


\paragraph{\ding{173} LLM-based equation synthesis.}

As \tasknameequation~assessing whether the LLMs can infer the correctness of equation (i.e., binary classification), for each human-written positive equation, we have to craft counterpart negative equations. To this end, for each positive equation, we prompt GPT-4 to synthesize a negative equation based on the paper context. We repeat this prompt (with a high decoding temperature) until three different negative equations are synthesized.\footnote{The number of negative equations is empirically decided.}


\paragraph{\ding{174} LLM-based filtering.}

However, the LLM-synthetic equations can be context-unaligned, i.e., some synthesized equations contain notation that is never defined in the paper context, which becomes a superficial shortcut and too effortless for LLMs to identify.
%
To improve data quality, we prompt GPT-4 to identify context-unaligned negative equations. We then eliminate the positive equation and its negative counterparts, where all three negative counterparts are unaligned. This filtering leads to a final of 1,449 positive equations and 4,347 negative equations (each positive equation has three negative counterparts, and at least one negative counterpart is ``challenging''). 



\paragraph{\ding{175} Expert-based examination.}

Furthermore, it's also possible that synthesized negative equations are actually correct (i.e., false negative) --- even if the negative and positive equations are written differently, the final compiled results might be the same.
We then employ human experts to review the data further and filter out false negative equations, checking the classification instances for accuracy.
%

We asked 5 senior PhD students who are experienced in AI research to check all instances. We ask human experts to consider the following criteria for each positive equation and its negative counterparts (each pair): i) \textbf{Are all equations grammatically correct?} ii) \textbf{After compilation, are all negative equations different from the positive ones?} We ask every human expert to use external LaTeX compilation tools (e.g., TeXlive), and identify the pairs that cannot meet the criteria. Each pair is examined by at least two experts, and we only keep pairs that all experts decide to keep. After this strict examination, a total of 1,049 pairs are eventually kept (27.6\% pairs are filtered)


\paragraph{Final data.}
We finally obtain 1,049 positive equations (each has three negative counterparts).
We show data statistics of ~\tasknameequation~in Table~\ref{tab:equations_statistics} and data examples in Figure~\ref{fig:case_eq}.


\subsection{\tasknameexperimentlong}
\label{subsec:data_collection_exp}

Given a research topic, such as a novel ML algorithm, a qualified researcher can design a solid experiment plan for it, and clarify underlying motivation to ensure the reliability of the designed experiment. Unlike the concurrent works that focus on the experiment implementation~\citep{lu2024aiscientist,huang2024mlagentbench}, we emphasize the importance of assessing the high-level experiment design of LLMs before the subsequent implementation to avoid any expensive execution iteration.
Therefore, as shown in Figure~\ref{fig:intro_three_tasks}, we formulate \textbf{\ColoredEXP}~as a text-generation task that takes pre-experiment paper context as input, and then generates the experiment and explanation list.


\paragraph{\ding{172} Data crawling.}

As for the data source, we first collect $\geq$ 10k papers' data from arXiv, including LaTeX sources and PDFs, which cover broad AI categories, including \texttt{cs.AI}, \texttt{cs.CL}, and \texttt{cs.CV}, from year \texttt{2018} to \texttt{2023}. Similarly, to ensure the source data quality, we only use papers that have appeared at well-known conferences.

\paragraph{\ding{173} Domain-expert annotation.}

Making a reliable and executable experiment plan requires solid foundation knowledge of a specific research area. Consequently, we set a high standard for choosing annotators: i) be a senior Ph.D. student with at least one peer-reviewed publication in leading AI venues; ii) have more than 4 years of AI research experience; iii) frequently serve as conference reviewers. Finally, we invite a total of 10 qualified experts to participate in our data collection procedure. Given the 10k crawled papers, we first ask every annotator to bid on the papers that they are interested in. After bidding, each of them is assigned 10 papers, i.e., a total of 100 papers to be annotated. 
During annotation, we post each paper PDF on online Google Drive and ask the annotator to first carefully read the whole paper. Then, we ask them to identify and locate the key experiments in each paper (i.e., highlighting the relevant paragraphs of each experiment). We don't consider some trivial experiments, such as those supplemental analyses in the appendix section.
For each identified experiment, the annotator has to concisely answer two questions: i) \textbf{What did this experiment do?} ii) \textbf{Why did the paper authors conduct this experiment?} 
In other words, we ask the annotator to summarize all the key experiments in this paper and explain the underlying motivations based on their rich domain experience. 


\paragraph{\ding{174} Multi-round peer discussion.}

Intuitively, different experts might have different opinions on the same research topic. Particularly, when explaining the underlying motivation of an experiment, adopting only a single expert's opinion might introduce bias to our annotation.
Hence, we conduct a further multi-round peer discussion. For each paper, where all the key experiments are identified, summarized, and explained, we ask a different expert (reviewer) to review the annotation by considering the following three criteria: i) \textbf{Are the identified experiments all the key experiments?} ii) \textbf{Does each experiment summarization covers all key information?} iii) \textbf{Does each explanation sound reasonable and reliable?} 
Each reviewer must leave comments on the online PDF regarding the above criteria, and then the annotator must respond to each comment --- either accept the suggestion and revise the previous annotation or provide a ``rebuttal'' to the reviewer to uphold the annotation. This discussion is iterative until both opinions align. 
Eventually, for each paper, we collect two lists: i) the experiment list, summarizing each experiment step of the paper; ii) the explanation list, the underlying motivations that are one-one corresponding to the experiment.
%

\paragraph{Final data.}

After annotation, we use the pre-experiment context of each paper (according to the first-experiment location identified by the annotator) as the input. Furthermore, we use GPT-4 to delete any sentence that potentially leaks the experiment from the input.\footnote{About 9.8\% sentences are deleted.}
Similar to the~\tasknameequation, we utilize the source LaTeX as the input text to avoid PDF paring noise.
As for the image input, we collect those figures within each paper's source LaTeX package and only keep figures that are used in the pre-experiment context.
Overall, a total of 100 instances are collected. As shown in Figure~\ref{fig:intro_three_tasks}, the input of each instance is the pre-experiment context (including the figures), and the ground-truth output is the expert-annotated experiment plan and the explanations. Table~\ref{tab:experiment_statistics} shows data statistics and Figure~\ref{fig:case_exp} illustrates the sample case in \tasknameexperiment.



\subsection{\tasknamereviewlong}
\label{subsec:data_collection_review}

Another critical research task is paper review. Previous works have demonstrated the usefulness of the LLM-based review feedback~\citep{gao2024reviewer2,jin2024agentreview,lu2024aiscientist}. However, as indicated by~\citet{du2024llms,liang2024can}, LLMs only excel at summarizing the research strengths while falling significantly short on weakness criticism. 
Hence, we build \textbf{\ColoredReview}~for particularly investigating the LLM-generated weaknesses.

\paragraph{\ding{172} Data crawling.}
We first crawl a total of 3,779 anonymous submissions of \textit{ICLR 2023} from OpenReview,\footnote{We adopt ICLR because it releases full submissions, while some other conferences only release accepted papers.} including PDF and other meta information (e.g., scores, decisions, and tracks). As the \textit{ICLR 2023} has 13 distinct tracks while the paper distribution across different tracks is highly biased, we then uniformly sample papers from different research tracks to improve the domain diversity. Meanwhile, during sampling, we also keep the accept/reject papers distributed equally to avoid data bias. In a word, we finally collect a total of 1,000 papers (500 accepted; 500 rejected), uniformly covering all 13 tracks. Please refer to Figure~\ref{fig:task3_data_distribution} for the track and score distribution of the 1,000 papers.

\paragraph{\ding{173} Extraction of human-written weaknesses.}

Since the raw comments crawled from \textit{ICLR 2023} are mixed with both strengths and weaknesses, we further employ GPT-4 to extract all the weaknesses from each reviewer's comments and compose multiple weaknesses into a list. Notably, we force GPT-4 to keep the original text of the reviewer, i.e., all weaknesses in our dataset are those original sentences written by the reviewer without any modifications.\footnote{We manually checked GPT-4's extraction results of 200 cases --- GPT-4 only missed \(\leq\)1\% of reviewer-written weaknesses and maintained almost all the original text.} What's more, sometimes one reviewer might repeatedly mention the same weakness throughout the comment. In this case, we simply keep all the repeated weaknesses because, if one weakness is repeatedly mentioned by the reviewer, it's intuitively an important weakness that the reviewer wants to emphasise; accordingly, keeping the repeat items can penalize LLMs more on missing this weakness.

For each paper, we can finally get multiple weakness lists (one weakness list per reviewer, one paper can have multiple reviewers). We further delete a few papers without any weaknesses found in the raw comments, resulting in a total of 993 instances, i.e., 993 \{paper, weakness lists\} pairs.

\paragraph{\ding{174} Input data processing.}

As we mentioned before, we crawl papers from OpenReview instead of arXiv because the under-review paper draft is required for this task. However, not every paper from OpenReview can be found on arXiv, i.e., the source LaTeX code and figures of most under-review papers are unavailable. Therefore, 
we utilize VILA~\citep{lin2023vila}~to parse text data out from the PDF; we also employ PDFFigures-2.0~\citep{clark2016pdffigures} to extract all the figures and tables (in image) from the paper, as Vila is not good at processing the table data.

\paragraph{Final data.}
Our final data is composed of 993 instances, each input is paper text along with figure/table images, and each output is peer reviewers' weakness lists. Table~\ref{tab:review_weakness_statistics} shows data statistics; Figure~\ref{fig:case_weakness} presents an example of the data instances. We show the data diversity (score and track distribution) in Figure~\ref{fig:task3_data_distribution}.

\subsection{\tasknamemetareviewlong}
\label{subsec:data_collection_metareview}

In addition to identifying weaknesses in paper drafts, a more challenging research task that requires more senior research experience is conducting meta-reviewing. Given a paper submission, along with individual reviews and author rebuttals, meta-reviewing is not to summarize individual reviews. Instead, a meta-reviewer must go through all the information and make a final recommendation. This requires the meta-reviewer to identify deficient/unreliable review segments (e.g., if a viewpoint is too subjective or contains factual errors) in each individual review and make a decision based on the non-deficient ones. This task demands years of experience in the relevant domain; even for human experts, only senior researchers are typically qualified for meta-reviewing. Therefore, as illustrated in Figure~\ref{fig:intro_three_tasks}, we also investigate how LLMs assist meta-reviewers, specifically in identifying deficient review points.

We reuse the \textbf{\ColoredMetaReview} dataset from our recent work~\citep{du2024llms}, where we crawled papers' initial submissions along with their reviews from OpenReview and employed more than 40 AI research experts to label each review segment (i.e., deficient or not), with detailed human explanations. In total, there were 100 papers with 380 human reviews. Each review was divided into sentence-level segments, resulting in 11,376 review segments (viewpoints).




\section{Evaluation Criteria}
\label{sec:eval_metrics}

For~\textbf{\ColoredEQ}, we adopt F$_1$ as the classification criterion.
For~\tasknameexperiment~and~\tasknamereview, since both tasks have free-form outputs, we develop several novel task-specific metrics in addition to the conventional ROUGE~\citep{lin2004rouge}. 


We use LLMs to evaluate the experiment list of \textbf{\ColoredEXP}. Specifically, given a model-predicted experiment list $p$, and the ground-truth list $g$, we calculate:

{\small
\setlength{\abovedisplayskip}{0pt}
\setlength{\belowdisplayskip}{3pt}
\begin{align}
\textrm{\MetricExpPrec} &= \frac{1}{m} \sum_{i=1}^{m} f(p_i, g) 
\label{eq:metric_exp_prec} \\
%
\textrm{\MetricExpRecall} &= \frac{1}{n} \sum_{j=1}^{n} f(g_j, p) 
\label{eq:metric_exp_recall}
\end{align}
}
where the $m$ and $n$ are the list length of $p$ and $g$; $f(.)$ represents the LLM prompting, where we prompt LLM to decide whether each predicted experiment item ($p_i$) is entailed by the whole ground-truth list ($g$), proceeding with binary output, and vice versa. Intuitively, \MetricExpPrec~reflects how many prediction experiments match ground-truth experiments. In this work, we used GPT-4o as an evaluator.

While for the explanation generation of \tasknameexperiment, as the prediction experiments are one-on-one corresponding to the ground truth, we adopt a semantic-based metric:
{\small
\setlength{\abovedisplayskip}{5pt}
\setlength{\belowdisplayskip}{5pt}
\begin{align}
    \textrm{\MetricExpMatch} = \frac{1}{m} \sum_{i=1}^{m} \textrm{sim}(p_i, g_i) 
    \label{eq:metric_exp_match}
\end{align}
}
where we use SentenceBERT~\citep{reimers2019sentence} to measure the semantic similarity between $p_i$ and $g_j$.



Unlike \tasknameexperiment, the ground truth of \textbf{\ColoredReview}~is multiple reviewers' weakness lists. Instead of merely merging the opinions of various reviewers into one flattened list and keeping LLM-as-judge as the metric (which is not only costly but also loses the structural information of diverse research perspectives), we employ the following semantic-based metric to efficiently evaluate predicted weaknesses:

{\small
\setlength{\abovedisplayskip}{0pt}
\setlength{\belowdisplayskip}{3pt}
\begin{align}
    \textrm{\MetricReviewPrec} &= \frac{1}{m} \sum_{i=1}^{m} \left(\frac{1}{r} \sum_{k=1}^{r} \max_{j} \textrm{sim}(p_i, g^{k}_{j})\right) 
    \label{eq:metric_review_prec}
    \\
    \textrm{\MetricReviewRecall} &= \frac{1}{r} \sum_{k=1}^{r} \left(\frac{1}{n_k} \sum_{j=1}^{n_k} \max_{i} \textrm{sim}(g^{k}_{j}, p_i)\right) 
    \label{eq:metric_review_recall}
\end{align}
}
where $r$ is the number of reviewers of the given paper, $n_k$ means the length of $k$-th reviewer's weakness list, and $g^k_j$ indicates the $j$-th item in $k$-th reviewer's weakness list. 

Additionally, in the real world, 
we would think a review weakness is reliable if it is specific to a paper. Meanwhile, we also hope the review is informative, i.e., no excessive similar weaknesses in one review. 
Inspired by the classic TF-IDF, we propose a novel review diversity metric:
{\small
\setlength{\abovedisplayskip}{0pt}
\setlength{\belowdisplayskip}{0pt}
\begin{align}
{\textrm{\MetricReviewIDF}}&=\frac{1}{w} \sum_{j=1}^w\left(\frac{1}{m_j} \sum_{i=1}^{m_j} \log \left(\frac{m_j}{O_i^j}\right) \times \log \left(\frac{w}{R_i^j}\right)\right) 
\label{eq:metric_review_idf}
\\
O_i^j&=\sum_{k=1}^{m_j} \textrm{sim}(p_i^j, p_k^j) 
\\
R_i^j&=\sum_{l=1}^w \max _{s} \textrm{sim}(p_i^j, p_s^l) 
\end{align}
}



where the $w$ is the total number of papers in the dataset, $p^j$ is $j$-th paper's prediction weakness list, $p^j_i$ is the $i$-th weakness in $p^j$. Moreover, ${O_i^j}$ calculates the intra-paper occurrence frequency of $p^j_i$; ${R_i^j}$ is the ``soft'' number of papers that also contain the $p^j_i$, which is computed by summing the maximum similarity scores between $p^j_i$ and other paper's weaknesses. In a word, ${O_i^j}$ measures informativeness, and ${R_i^j}$ measures specificity. The complete \MetricReviewIDF~consider both aspects and reflects the overall weakness diversity.

For \textbf{\ColoredMetaReview}, we use F$_1$ score as the classification metric; while for the deficiency explanation, we use ROUGE~\citep{lin2004rouge} and BERTScore \citep{DBLPZhangKWWA20} to reflect how well the model-generated explanation aligns with the expert's annotation.





\section{Experiments and Analyses}

In this section, we conduct extensive experiments on~\dataname, across various mainstream LLMs, to quantify the current LLMs' capacity to tackle high-level research tasks. Specifically, \cref{subsec:experiment_equation} for \textbf{\ColoredEQ}, \cref{subsec:experiment_exp} for \textbf{\ColoredEXP}, \cref{subsec:experiment_review} for \textbf{\ColoredReview};
and \cref{subsec:experiment_meta_review} for \textbf{\ColoredMetaReview}.
Please refer to Appendix~\ref{appendix:details_model} for running details of the LLMs.

\subsection{\tasknameequationlong}
\label{subsec:experiment_equation}

\begin{table}[t!]
\centering
\caption{Various LLMs' performances on \protect\ColoredEQ~task (1,049 positive and 3,147 negative samples). ``All-positive'' indicates a baseline that predicts all equations as positive.}
\resizebox{0.48\textwidth}{!}{
\begin{tabular}{lrrr}

\toprule

\textbf{Methods} 
& \multicolumn{1}{c}{\textbf{F$_{1}$}} & \textbf{Prec.} & \textbf{Rec.} \\ 

\midrule

All-Positive         &
40.00  & 25.00 & 100.00                                    \\ 

\hline

\rowcolor[HTML]{E7E6E6} 
\multicolumn{4}{c}{\cellcolor[HTML]{E7E6E6}\textit{\textbf{Open-source LLMs}}}  \\ 

\hline

 
OLMo-7B~\citep{Groeneveld2023OLMo}             & \cellcolor[HTML]{E7E8F5}13.64	 & \cellcolor[HTML]{E7E8F5}11.93	&  \cellcolor[HTML]{E7E8F5}15.91           \\

Mistral-7B~\citep{jiang2023mistral}          & \cellcolor[HTML]{E7E8F5} 28.45&	\cellcolor[HTML]{E7E8F5}19.28	& \cellcolor[HTML]{E7E8F5}54.24           \\

Mixtral-8x22B-MoE~\citep{jiang2024mixtral}   & \cellcolor[HTML]{E7E8F5}40.90	& \cellcolor[HTML]{E7E8F5}26.15	& \cellcolor[HTML]{E7E8F5}93.80           \\

Qwen 2.5-72B~\citetalias{qwen2.5}        & \cellcolor[HTML]{E7E8F5} 31.22	& \cellcolor[HTML]{E7E8F5} 26.28	& \cellcolor[HTML]{E7E8F5} 57.40          \\

Llama 3.1-70B~\citep{Llama3.1:online}       & \cellcolor[HTML]{E7E8F5} 33.08	& \cellcolor[HTML]{E7E8F5}22.14	 & \cellcolor[HTML]{E7E8F5}65.39        \\

\hline

\rowcolor[HTML]{E7E6E6} 
\multicolumn{4}{c}{\cellcolor[HTML]{E7E6E6}\textit{\textbf{Closed-source LLMs}}} \\ 

\hline
 
Gemini 1.5 Pro~\citep{team2023gemini}      & \cellcolor[HTML]{F7D7D6} 46.74	& \cellcolor[HTML]{F7D7D6} 32.05	& \cellcolor[HTML]{F7D7D6} 86.27          \\

Claude 3.5 sonnet~\citep{Claude3.5Sonnet:online}   & \cellcolor[HTML]{F7D7D6} 45.13	& \cellcolor[HTML]{F7D7D6} 29.48	& \cellcolor[HTML]{F7D7D6} \textbf{96.18}  \\

GPT-4o~\citep{GPT4o:online}              & \cellcolor[HTML]{F7D7D6} 40.35	& \cellcolor[HTML]{F7D7D6} 30.79	& \cellcolor[HTML]{F7D7D6} 58.53           \\

o1-preview~\citep{OpenAI_o1:online}                  & \cellcolor[HTML]{F7D7D6} 46.35	& \cellcolor[HTML]{F7D7D6} 31.43	& \cellcolor[HTML]{F7D7D6} 88.27          \\

o3-mini~\citep{OpenAIo319:online}                  & \cellcolor[HTML]{F7D7D6} \textbf{47.98}	& \cellcolor[HTML]{F7D7D6} \textbf{34.34}	& \cellcolor[HTML]{F7D7D6} 79.59          \\

\bottomrule

\end{tabular}
}
\label{tab:equation_main_tab}
\end{table}

\paragraph{Settings.}
As different LLMs have distinct context windows, to ensure a fair comparison, we fix the maximum input length for all models. According to Table~\ref{tab:equations_statistics}, we empirically use 1,000 words for both contexts before and after equations, i.e., 2,000 surrounding words.

\paragraph{Main results.}
Table~\ref{tab:equation_main_tab} shows the main results. 
Firstly, a simple baseline that predicts all equations as positive achieves 40\% F$_1$ (due to the 1:3 of positive and negative equations), while nearly all open-source LLMs even cannot beat this naive baseline. Notably, though the performance of Mixtral is slightly superior to the baseline, the extremely biased precision and recall scores imply that Mixtral is also simply predicting almost all samples as positive instead of truly inferring. Meanwhile, compared to the All-Positive baseline, the performance superiority of the strong close-source LLMs is not significant, the best LLM on this task only obtains 47.98\%, which demonstrates the challenge of \tasknameequation~compared with other similar benchmarks~\citep{song2023nlpbench}. The generally high recall with low precision of all LLMs also indicates real-world risks, e.g., relying on LLMs to check the validity of equations in paper review.

\paragraph{$\mathcal{Q}$: Do more contexts boost performance?} \tasknameequation~places high demands on reasoning within the scientific context. To quantify the impact of input context length, we scale the input length (per side) from 100 to 1,500 words. As shown in Figure~\ref{fig:task1_equation_scaling}, for the open-source LLMs (Llama and Qwen), an appropriate context length can boost the performance; while for GPT-4o, scaling up the context length doesn't contribute much to the F$_1$. However, during the scaling, we find that the precision of GPT-4o is gradually increased, and the recall is decreased accordingly; considering the label distribution of \tasknameequation, we believe precision can better reflect the model's true capacities on this task. Thus, we anticipate that scaling up context shall be beneficial to those strong close-source LLMs such as GPT-4o.

\subsection{\tasknameexperimentlong}
\label{subsec:experiment_exp}

\begin{table*}[t]
\centering
\tiny
\caption{Various LLMs' performances on the 100 instances of \protect\ColoredEXP. The explanation generation is based on the oracle experiments to prevent error propagation. ``Copy Input'' directly copies each experiment idea as the explanation.}
\resizebox{0.96\textwidth}{!}{
\begin{tabular}{lrrrrrr}

\toprule

                                   & \multicolumn{3}{c}{\textbf{Experiment Design}}     
                                   & \multicolumn{3}{c}{\textbf{Experiment Explanation}}  
                                   \\ 
                                   \cmidrule(r){2-4} \cmidrule(l){5-7}
\multirow{-2}{*}{\textbf{Methods}} & \multicolumn{1}{c}{\textbf{\MetricExpF}} & \multicolumn{1}{c}{\textbf{\MetricExpPrec}} & \multicolumn{1}{c}{\textbf{\MetricExpRecall}} & \multicolumn{1}{c}{\textbf{\MetricExpMatch}} & \multicolumn{1}{c}{\textbf{ROUGE-L}} & \multicolumn{1}{c}{\textbf{ROUGE-1}} \\ 

\midrule

Copy Input                         & ---	                                  & ---	                                        & ---	                                      & {40.32}	                                     & {22.06}	                                & {25.28}                                 \\ \hline
\rowcolor[HTML]{E7E6E6} 
\multicolumn{7}{c}{\cellcolor[HTML]{E7E6E6}\textit{\textbf{Open-source LLMs}}}                                                                                                                                                                                                           \\ \hline
\rowcolor[HTML]{E7E8F5} 
\cellcolor[HTML]{FFFFFF} OLMo-7B~\citep{Groeneveld2023OLMo}                            & 14.80	& 17.50	& 19.80                                   & 45.78                                   & 26.30                               & 30.38                               \\
\rowcolor[HTML]{E7E8F5} 
\cellcolor[HTML]{FFFFFF} Mistral-7B~\citep{jiang2023mistral}                         & 18.96	& 24.83	& 21.38                                    & 50.18                                   & \textbf{30.20}                      & 34.69                               \\
\rowcolor[HTML]{E7E8F5} 
\cellcolor[HTML]{FFFFFF} Mixtral-8x22B-MoE~\citep{jiang2024mixtral}                  & 23.16	 & 24.45	& 30.57                                    & 49.07                                   & 29.96                               & 34.53                               \\
\rowcolor[HTML]{E7E8F5} 
\cellcolor[HTML]{FFFFFF} Llama 3.1-70B~\citep{Llama3.1:online}                      & 22.92	& 23.10	& 29.76                                    & 50.05                                   & 29.33                               & 34.11                               \\
\rowcolor[HTML]{E7E8F5} 
\cellcolor[HTML]{FFFFFF} Qwen 2.5-72B~\citetalias{qwen2.5}                       & 24.28	& 22.48	 & 34.44                                    & 51.12                                   & 29.46                               & 34.68                               \\ \hline
\rowcolor[HTML]{E7E6E6} 
\multicolumn{7}{c}{\cellcolor[HTML]{E7E6E6}\textit{\textbf{Closed-source LLMs}}}                                                                                                                                                                                                          \\ \hline
\rowcolor[HTML]{F7D7D6} 
\cellcolor[HTML]{FFFFFF} Gemini 1.5 Pro~\citep{team2023gemini}                     & 27.25	& 28.66	 & 34.92                                    & 52.87                                   & 28.52                               & 33.80                               \\
\rowcolor[HTML]{F7D7D6} 
\cellcolor[HTML]{FFFFFF} Claude 3.5 sonnet~\citep{Claude3.5Sonnet:online}                  & 27.99	& 24.48	& \textbf{42.09}                                    & 53.03                                   & 18.75                               & 26.15                               \\
\rowcolor[HTML]{F7D7D6} 
\cellcolor[HTML]{FFFFFF} GPT-4o~\citep{GPT4o:online}                             & 25.03	& 22.25	 & 36.59                           & 54.79                                   & 27.54                               & 34.31                               \\
\rowcolor[HTML]{F7D7D6} 
\cellcolor[HTML]{FFFFFF} o1-preview~\citep{OpenAI_o1:online}                         & 30.13	& 28.13	 & 38.59                                    & \textbf{58.55}                          & 29.11                               & \textbf{36.70}                      \\ 

\rowcolor[HTML]{F7D7D6} 
\cellcolor[HTML]{FFFFFF} o3-mini~\citep{OpenAIo319:online}                         & \textbf{30.17}	& \textbf{28.70}	 & 37.67                                    & 54.01                          & 20.71                               & 29.14                     \\ 

\bottomrule

\end{tabular}

}
\label{tab:experiment_explanation_main_tab}
\end{table*}

\paragraph{Settings.}

Similarly, we unify the input context length of different LLMs to ensure a fair comparison. According to Table~\ref{tab:experiment_statistics}, we set 2,000 and 3,000 input words for open- and closed-source LLMs, respectively. Meanwhile, as experiment explanation is the subsequent task of experiment design, using model-generated experiments can propagate errors in explanation, leading to inferior results for most LLMs. To this end, we provide LLMs with the oracle experiments when generating explanations.

\paragraph{Main results.}
Table~\ref{tab:experiment_explanation_main_tab} shows the main results. For the experiment design, the closed-source LLMs generally outperform open-source LLMs. However, the score values of all LLMs are relatively low (20\%$\sim$30\%), implying the LLMs consistently miss ground-truth experiments from the origin paper (low recall), and they tend to generate more novel experiments that didn't show in the origin paper (low precision).
As for the experiment explanation, the \MetricExpMatch~scores of closed-source LLMs still surpass the open-source LLMs.
Furthermore, there is a negative correlation between \MetricExpMatch~and ROUGE score, where the ROUGE scores of closed-source LLMs are broadly inferior. We find that the open-source LLMs often try to copy the terms or phrases from the given experiment, or even simply paraphrase the experiment instead of explaining, which results in a high superficial overlap with the ground-truth explanation.
This observation highlights the importance of adopting the proposed \MetricExpMatch~to avoid evaluation bias of traditional generation metrics. 

\paragraph{$\mathcal{Q}_1$: What is the quality of the model-generated novel experiments?}

The low \MetricExpPrec~of LLMs in Table~\ref{tab:experiment_explanation_main_tab} indicates the creativity of LLMs in generating novel experiments. We then randomly sample 15 papers from the \tasknameexperiment~and ask 3 experts to manually review the model-generated novel experiments. Specifically, we ask the experts to judge the necessity of the novel experiments, where we set three necessity levels: ``A'' indicates the experiment is necessary/mandatory to support the main claim, ``B'' represents optional/supplementary experiments, and ``C'' for those unrelated experiments (see Appendix~\ref{appendix:details_human_eval_experiment_necessity} for evaluation details). Table~\ref{tab:experiment_necessity_human_eval} shows the necessity scores of the three strongest LLMs. We find that LLMs consistently generate a lot of novel experiments, especially the Claude; though most of them are optional, even fancy/unrelated experiments, there are still a considerable amount of necessary experiments generated, e.g., the results of o1. We further find that some novel experiments can be regarded as useful supplementary analyses w.r.t. the human experiments. Table~\ref{tab:exp_necessity_cases} shows examples of model-suggested experiments.

\begin{table}[t!]
\centering
\tiny
\caption{The human evaluation results on the novel experiments suggested by LLMs. ``A'', ``B'', and ``C'' represent the different quality level (i.e., necessity); ``A'' is the best level.}
\resizebox{0.48\textwidth}{!}{

\begin{tabular}{lrrr}
\toprule
\multirow{2}{*}{\textbf{Models}} & \multicolumn{1}{c}{\multirow{2}{*}{\textbf{\# of novel EXP}}} & \multicolumn{2}{c}{\textbf{Necessity~(\%)}}                 \\ \cmidrule{3-4} 
                        & \multicolumn{1}{c}{}                                 & \multicolumn{1}{c}{A} & \multicolumn{1}{c}{B} \\ 
                        
                        \midrule
                        
Gemini 1.5 Pro          & 59                                                   & 30.59                 & 45.76                 \\
Claude 3.5 sonnet       & 112                                                  & 21.78                 & 50.00                 \\
o1-preview              & 71                                                   & 35.84                 & 36.61                 \\ 

\bottomrule
\end{tabular}

}

\label{tab:experiment_necessity_human_eval}
\end{table}

\paragraph{$\mathcal{Q}_2$: Can self-contained experiment design enhance the experiment explanation?}

\begin{table}[t!]
\centering
\tiny
\caption{The impact on \MetricExpMatch~scores of maintaining the experiment's self-containment for \protect\ColoredEXP.}
\resizebox{0.48\textwidth}{!}{
\begin{tabular}{lcr}
\toprule

\textbf{Models}            & \multicolumn{1}{c}{\textbf{One-by-One}} & \multicolumn{1}{c}{\textbf{Whole-List}} \\ 

\midrule

Llama 3.1-70B     & 50.05                          & 49.36~{\tiny (\MydownArrow{0.7})}                          \\
Qwen 2.5-72B      & 51.12                          & 48.56~{\tiny (\MydownArrow{2.6})}                          \\ 
\midrule
Gemini 1.5 Pro    & 52.87                          & 57.48~{\tiny (\MyUpArrow{4.6})}                         \\
Claude 3.5 sonnet & 53.03                          & 59.11~{\tiny (\MyUpArrow{6.1})}                          \\
GPT-4             & 55.03                          & 56.95~{\tiny (\MyUpArrow{1.9})}                          \\
GPT-4o            & 54.79                          & 58.54~{\tiny (\MyUpArrow{3.8})}                          \\
o1-preview                & 58.55                          & 61.58~{\tiny (\MyUpArrow{3.0})}                          \\ 
\bottomrule

\end{tabular}
}

\label{tab:experiment_two_prompts}
\end{table}

When generating the explanation in Table~\ref{tab:experiment_explanation_main_tab}, we provide LLMs with each individual experiment and let them explain one by one, because we find that, when providing the whole experiment list, those open-source models only explain partial experiments because of their poor instruction-following capacity. However, there are intuitively some semantic or logical relations between different experiments, e.g., some experiments are prerequisites to others.
Therefore, this one-by-one prompting might break the self-containment of an experiment plan. Consequently, we test with the ``whole-list'' prompting, where the LLMs are given the complete experiment list and are asked to explain all experiment steps together.

As shown in Table~\ref{tab:experiment_two_prompts}, unlike the open-source LLMs, the explanation performances of those closed-source LLMs are generally improved after adopting whole-list prompting. According to further manual checking, after maintaining the self-containment of the experiments, the LLMs can refer to other experiments and better grasp the underlying motivation of the current experiment.

\paragraph{$\mathcal{Q}_3$: Do human evaluation results align with automatic metrics for explanation?}

\begin{table}[t!]
\centering
\tiny
\caption{The human evaluation results on LLMs' output explanations of \protect\ColoredEXP. ``Acc. ratio'' means how many model outputs are accepted by the annotator.}

\resizebox{0.33\textwidth}{!}{
\begin{tabular}{lr}
\toprule
\rowcolor[HTML]{FFFFFF} 
\textbf{Models}            & \multicolumn{1}{c}{\cellcolor[HTML]{FFFFFF}\textbf{Acc. ratio}} \\ \midrule
Llama 3.1-70B     & 22.93                                                  \\
Gemini 1.5 Pro    & 55.07                                                  \\
Claude 3.5 sonnet & 61.46                                                  \\
GPT-4o            & 69.72                                                  \\
o1-preview                & \textbf{76.14}                                                  \\ \bottomrule
\end{tabular}
}
\label{tab:experiment_human_eval}
\end{table}


As the explanation can be open-ended, in this paragraph, we provide the human evaluation results on different LLMs' experiment explanation outputs. In detail, we randomly select 20 out of 100 papers and ask 5 annotators to read the experiments along with each model's explanations; we then let the annotator decide whether each model's explanation is acceptable (see Appendix~\ref{appendix:details_human_eval} for more details). Table~\ref{tab:experiment_human_eval} illustrates the results, where the score variance is higher than Table~\ref{tab:experiment_explanation_main_tab}. However, the performance ranking of both tables is perfectly correlated with each other (Spearman's rank correlation coefficient $=$ 1), demonstrating the effectiveness of \MetricExpMatch.

\begin{table*}[t]
\centering
\tiny
\caption{Various LLMs' performances on the 993 instances of \protect\ColoredReview.}
\resizebox{\textwidth}{!}{

\begin{tabular}{lrrrr}
\toprule

\rowcolor[HTML]{FFFFFF} 
\cellcolor[HTML]{FFFFFF}                                   & \multicolumn{1}{c}{\cellcolor[HTML]{FFFFFF}}                                   & \multicolumn{1}{c}{\cellcolor[HTML]{FFFFFF}}                                          & \multicolumn{1}{c}{\cellcolor[HTML]{FFFFFF}}                                       & \multicolumn{1}{c}{\cellcolor[HTML]{FFFFFF}\textbf{Weakness Diversity}} \\ \cmidrule{5-5}
\rowcolor[HTML]{FFFFFF} 
\multirow{-2}{*}{\cellcolor[HTML]{FFFFFF}\textbf{Methods}} & \multicolumn{1}{c}{\multirow{-2}{*}{\cellcolor[HTML]{FFFFFF}\textbf{\MetricReviewF~(\%)}}} & \multicolumn{1}{c}{\multirow{-2}{*}{\cellcolor[HTML]{FFFFFF}\textbf{\MetricReviewPrec~(\%)}}} & \multicolumn{1}{c}{\multirow{-2}{*}{\cellcolor[HTML]{FFFFFF}\textbf{\MetricReviewRecall~(\%)}}} & \multicolumn{1}{c}{\cellcolor[HTML]{FFFFFF}\textbf{\MetricReviewIDF~($\uparrow$)}}    \\ 

\midrule

Human Review                                                & ---                                                                              & ---                                                                                     & ---                                                                                  & { 7.69}                                                           \\ \hline
\rowcolor[HTML]{E7E6E6} 
\multicolumn{5}{c}{\cellcolor[HTML]{E7E6E6}\textit{\textbf{Open-source LLMs}}}                                                                                                                                                                                                                                                                                                                  \\ \hline
OLMo-7B~\citep{Groeneveld2023OLMo}                                                    & \cellcolor[HTML]{E7E8F5}43.25                                                  & \cellcolor[HTML]{E7E8F5}40.38                                                         & \cellcolor[HTML]{E7E8F5}47.04                                                      & \cellcolor[HTML]{E7E8F5}2.45                                         \\
Mistral-7B~\citep{jiang2023mistral}                                                 & \cellcolor[HTML]{E7E8F5}42.03                                                  & \cellcolor[HTML]{E7E8F5}43.80                                                         & \cellcolor[HTML]{E7E8F5}40.77                                                      & \cellcolor[HTML]{E7E8F5}1.17                                         \\
Mixtral-8x22B-MoE~\citep{jiang2024mixtral}                                          & \cellcolor[HTML]{E7E8F5}43.23                                                  & \cellcolor[HTML]{E7E8F5}\textbf{44.59}                                                & \cellcolor[HTML]{E7E8F5}42.23                                                      & \cellcolor[HTML]{E7E8F5}0.98                                         \\
Llama 3.1-70B~\citep{Llama3.1:online}                                              & \cellcolor[HTML]{E7E8F5}42.78                                                  & \cellcolor[HTML]{E7E8F5}43.19                                                         & \cellcolor[HTML]{E7E8F5}42.70                                                      & \cellcolor[HTML]{E7E8F5}2.60                                         \\
Qwen 2.5-72B~\citetalias{qwen2.5}                                               & \cellcolor[HTML]{E7E8F5}42.74                                                  & \cellcolor[HTML]{E7E8F5}43.80                                                         & \cellcolor[HTML]{E7E8F5}42.05                                                      & \cellcolor[HTML]{E7E8F5}1.21                                         \\ \hline
\rowcolor[HTML]{E7E6E6} 
\multicolumn{5}{c}{\cellcolor[HTML]{E7E6E6}\textit{\textbf{Closed-source LLMs}}}                                                                                                                                                                                                                                                                                                                 \\ \hline
Gemini 1.5 Pro~\citep{team2023gemini}                                             & \cellcolor[HTML]{F7D7D6}\textbf{48.75}                                         & \cellcolor[HTML]{F7D7D6}43.97                                                         & \cellcolor[HTML]{F7D7D6}55.08                                                      & \cellcolor[HTML]{F7D7D6}5.88                                         \\
Claude 3.5 sonnet~\citep{Claude3.5Sonnet:online}                                          & \cellcolor[HTML]{F7D7D6}47.85                                                  & \cellcolor[HTML]{F7D7D6}41.97                                                         & \cellcolor[HTML]{F7D7D6}56.00                                                      & \cellcolor[HTML]{F7D7D6}3.91                                         \\
GPT-4o~\citep{GPT4o:online}                                                     & \cellcolor[HTML]{F7D7D6}47.73                                                  & \cellcolor[HTML]{F7D7D6}42.09                                                         & \cellcolor[HTML]{F7D7D6}55.48                                                      & \cellcolor[HTML]{F7D7D6}\textbf{5.95}                                \\
o1-preview~\citep{OpenAI_o1:online}                                                 & \cellcolor[HTML]{F7D7D6}48.62                                                  & \cellcolor[HTML]{F7D7D6}42.54                                                         & \cellcolor[HTML]{F7D7D6}\textbf{57.08}                                             & \cellcolor[HTML]{F7D7D6}5.63                                         \\

o3-mini~\citep{OpenAIo319:online}                                                 & \cellcolor[HTML]{F7D7D6}46.33                                                  & \cellcolor[HTML]{F7D7D6}42.00                                                         & \cellcolor[HTML]{F7D7D6}51.99                                             & \cellcolor[HTML]{F7D7D6}5.85                                         \\

\hline
\rowcolor[HTML]{E7E6E6} 
\multicolumn{5}{c}{\cellcolor[HTML]{E7E6E6}\textit{\textbf{LLM Agent Framework}}}                                                                                                                                                                                                                                                                                                               \\ \hline
\AISci~(GPT-4o)~\citep{lu2024aiscientist}                                            & \cellcolor[HTML]{f0e4be}45.05                                                  & \cellcolor[HTML]{f0e4be}40.02                                                         & \cellcolor[HTML]{f0e4be}51.91                                                      & \cellcolor[HTML]{f0e4be}2.23                                         \\ 

\bottomrule

\end{tabular}

}
\label{tab:review_main_tab}
\end{table*}

\paragraph{$\mathcal{Q}_4$: Do more contexts boost performance?}

We also investigate the impact of input context length for~\tasknameexperiment. As shown in Figure~\ref{fig:task2_context_scaling}, we scale up the input pre-experiment context length from 0.1k to 10k tokens (10k is the length of the longest paper). For the experiment design, more input context does improve the performance of different LLMs, while this benefit stops after exceeding 8k tokens, which means that after the necessary information has been covered, scaling context becomes inefficient. 
Meanwhile, the explanation generation results reveal that LLMs primarily depend on given experiments rather than paper context to explain motivations. However, we do not expect this as we hope LLMs can explain the motivation based on a thorough understanding of the paper, just like how human experts do. Hence, there is still a considerable gap between the LLMs and humans in terms of grasping research motivations.

\paragraph{$\mathcal{Q}_5$: Does multi-modal input boost performance?}

Intuitively, besides the text, when designing experiments for a given research topic, the figures can provide rich supplementary information, such as an algorithm illustration that can help better understand this research topic and underlying motivations. Hence, we test the performance of different LMMs (Large Multimodal Models), including GPT4-o and InternVL2~\citep{chen2024far}. Table~\ref{tab:experiment_multi_modal}~shows the ablation results on the figure data. To our surprise, the figure data doesn't improve the LMMs' results in this task, even harming the performances.
 %
This might be due to the low informativeness of the figures, as figures usually consume more input tokens but act only as supplementary information to the text, indicating future work on developing LMMs that can effectively leverage the scientific figures.

\subsection{\tasknamereviewlong}
\label{subsec:experiment_review}

\paragraph{Settings.}
Intuitively, full paper content is necessary for paper reviewing. Therefore, instead of setting a maximum input length, in~\tasknamereview, we try to utilize the whole paper. As the input length of \tasknamereview~is extremely long (see Table~\ref{tab:review_weakness_statistics}), we adopt a ``split-combine'' method --- we first split the whole paper into smaller pieces and let LLMs predict the weaknesses of each piece separately; after that, we merge all pieces' weaknesses as a final prediction. For the length of each small piece, we set 2,000 and 3,000 words for open- and closed-source LLMs, respectively. Additionally, in this task, we also examine the performance of \AISci~\citep{lu2024aiscientist}, which enhances LLMs' paper review ability by leveraging advanced prompting techniques, e.g., self-reflection~\citep{shinn2024reflexion} and response ensembling~\citep{wang2022selfcons}.\footnote{We don't run \AISci~on \tasknameexperiment, because \AISci~takes model-generated ideas as the inputs, which are incompatible with our task setting.}

\paragraph{Main results.}

Table~\ref{tab:review_main_tab} shows the main results, where the closed-source LLMs' overall performances are generally superior to the results of open-source LLMs.
Similarly, closed-source LLMs are particularly excellent in \MetricReviewRecall~because of more generated weaknesses.
However, there is still a considerable gap in the weakness diversity between the LLMs and human experts.\footnote{The human's \MetricReviewIDF~score in Table~\ref{tab:review_main_tab} can be slightly underestimated. This is because we keep the repeated weaknesses in the human review, which affects the human review's informativeness (lower ITF) but is useful when calculating the \MetricReviewRecall~for LLMs.} Compared with human review, most LLM-generated weaknesses are vague and lack the necessary knowledge about some frontier research works.
Surprisingly, \AISci~performs worse than backbone GPT-4o, especially on \MetricReviewIDF, which suggests the challenge of \tasknamereview, i.e., simply adopting popular prompting techniques cannot well address this task.
%

\paragraph{$\mathcal{Q}_1$: Is the split-combine effective?}

Ideally, if the LLM has a sufficient context window size, splitting the input papers for separate processing is unnecessary. Consequently, in this paragraph,
we utilize the LLMs accepting long context input to compare ``split-combine'' with ``no-split'', i.e., letting LLMs write weaknesses by giving the full paper. In practice, we set the maximum number of input words to 20k, which ensures $\geq$95\% papers in the \tasknamereview~can be fully processed.
As shown in Table~\ref{tab:review_context}, compared with giving the full paper contexts, split-combine generally brings about superior performances. During manual checking, we find that, when full paper is available, LLMs frequently neglect some important sections and omit weaknesses accordingly, while split-combine ensures that the LLMs can carefully brainstorm weaknesses within each smaller piece. Surprisingly, the LLMs' performances with full paper context can be even worse than just remaining the first 3,000 words. This implies that even the current powerful long-context LLMs still fall short when processing long scientific documents.

\paragraph{$\mathcal{Q}_2$: Does multi-modal input boost performance?}

Our dataset covers both tables and figure illustrations extracted from the paper PDF as inputs. Intuitively, when reviewing a paper, both figures and tables are critical, not only for a better understanding, but also because some weaknesses are related to tables/figures.\footnote{We find that there is approximately one human-written weakness related to figures or tables in each paper.} Therefore, in Table~\ref{tab:review_multi_modal}, we adopt two LMMs to investigate the effectiveness of image inputs. Overall, image information, including both figures and tables, doesn't bring significant performance improvement, i.e., only InternVL2 gains a performance boost after incorporating figures; while tables slightly drop both models' results. This is probably because the LMMs cannot reason well over the information-intensive images, especially the table images.

\subsection{\tasknamemetareviewlong}
\label{subsec:experiment_meta_review}

\paragraph{Settings.} 
As individual review comments are split into multiple smaller segments (sentences), in order to avoid the performance variance that comes from the prompting, we follow \citet{du2024llms} to utilize two prompting strategies. i) \textbf{Labeling-All}: given everything necessary including a list of indexed review segments, require the LLM to output a list of triples, like \{id, reliable or not, explanation\}. ii)
\textbf{Select-Deficient}: Given everything necessary
including a list of indexed review segments, require the LLM to output a list of tuples, \{id, explanation\}, when it believes the ``id'' corresponds to a deficient segment.

To further enhance evaluation robustness, we ensemble the results obtained from the two prompting strategies using two methods. i) \textbf{Both ``No''}: if both
prompts classify a segment as deficient, we consider it to be deficient. ii) \textbf{Either ``No''}: if either
of the prompts labels a segment as Deficient, we consider it to be deficient.

\paragraph{Main results.}

\begin{table*}[t!]
    \centering
    \caption{From \citep{du2024llms}, various LLMs' performances on the 11,376 instances of \protect\ColoredMetaReview. The best F1 score among different prompt methods for a single model is \underline{underlined}. The best F1 score across all models is also \textbf{\underline{bold}}.}
    \resizebox{\linewidth}{!}{
    \begin{tabular}{@{}lrrrr}
        \toprule
        \multirow{2}{*}{\textbf{Models}} & \multicolumn{4}{c}{\textbf{Precision / Recall / F$_1$}} \\
        \cmidrule(lr){2-5}
        & \multicolumn{1}{c}{\textbf{Labeling-All}} & \textbf{Select-Deficient} & \multicolumn{1}{c}{\textbf{Both ``No''}} & \multicolumn{1}{c}{\textbf{Either ``No''}} \\

     \hline
    \rowcolor[HTML]{E7E6E6} 
    \multicolumn{5}{c}{\cellcolor[HTML]{E7E6E6}\textit{\textbf{Open-source LLMs}}}   \\ \hline
       
        Llama3-8B~\citep{llama3modelcard} & \cellcolor[HTML]{E7E8F5}7.73 / 45.95 / 12.22 
        & \cellcolor[HTML]{E7E8F5}11.47 / 30.29 / \underline{14.88} 
        & \cellcolor[HTML]{E7E8F5}11.37 / 21.27 / 12.46 
        & \cellcolor[HTML]{E7E8F5}8.19 / 53.61 / 13.35 \\
        Llama3-70B~\citep{llama3modelcard} & \cellcolor[HTML]{E7E8F5}13.63 / 42.49 / 18.19 
        & \cellcolor[HTML]{E7E8F5}13.95 / 31.16 / 17.46 
        & \cellcolor[HTML]{E7E8F5}16.16 / 23.51 / 16.67 
        & \cellcolor[HTML]{E7E8F5}12.46 / 50.02 / \underline{18.43} \\
        Qwen2-72B~\citep{bai2023qwen} & \cellcolor[HTML]{E7E8F5}9.97 / 26.60 / 12.96 
        & \cellcolor[HTML]{E7E8F5}11.35 / 34.61 / 14.64 
        & \cellcolor[HTML]{E7E8F5}9.07 / 15.13 / 9.62 
        & \cellcolor[HTML]{E7E8F5}10.49 / 43.00 / \underline{15.16} 
        \\

    \hline
    \rowcolor[HTML]{E7E6E6} 
    \multicolumn{5}{c}{\cellcolor[HTML]{E7E6E6}\textit{\textbf{Closed-source LLMs}}} \\ \hline

        Gemini 1.5~\citep{team2023gemini} & \cellcolor[HTML]{F7D7D6}16.58 / 34.13 / 19.76 
        & \cellcolor[HTML]{F7D7D6}14.71 / 43.60 / 19.72 
        & \cellcolor[HTML]{F7D7D6}17.01 / 27.05 / 18.28 
        & \cellcolor[HTML]{F7D7D6}14.46 / 50.37 / \underline{20.34} \\
         GPT-4~\citep{gpt4} & \cellcolor[HTML]{F7D7D6}14.91 / 34.49 / 18.38 
        & \cellcolor[HTML]{F7D7D6}17.18 / 34.59 / 20.30 
        & \cellcolor[HTML]{F7D7D6}18.71 / 21.40 / 16.85 
        & \cellcolor[HTML]{F7D7D6}14.72 / 47.68 / \underline{20.66} \\
        Claude Opus~\citep{claude_blog} & \cellcolor[HTML]{F7D7D6}16.86 / 34.26 / 20.35 
        & \cellcolor[HTML]{F7D7D6}17.69 / 26.61 / 18.71 
        & \cellcolor[HTML]{F7D7D6}17.14 / 18.70 / 15.78 
        & \cellcolor[HTML]{F7D7D6}16.94 / 42.12 / \textbf{\underline{21.99}}
        \\

        \bottomrule
    \end{tabular}
    }
    
    \label{tab:meta_review_main}
\end{table*}

We put the results of \citet{du2024llms} in Table~\ref{tab:meta_review_main}. Closed-source models (GPT-4, Claude Opus, and Gemini 1.5) generally outperform open-source models (Llama3-8B and 70B, Qwen2-72B) in F$_1$ score. Claude Opus achieves the highest F$_1$ scores, with GPT-4 and Gemini 1.5 performing slightly worse. Notably, recall scores are consistently higher than precision scores across all LLMs and prompting strategies, suggesting that LLMs tend to incorrectly identify segments as deficient.
Despite the superior performance of the closed-source models, their F$_1$ scores remain relatively
low even with different prompt strategies, highlighting the challenges LLMs face in such expertise-intensive tasks and emphasizing the importance of human expertise in the meta-reviewing process.

\paragraph{$\mathcal{Q}$: How about the LLMs explanation quality regarding the deficient review?}

As we also prompt the LLMs to generate the corresponding explanation on why they think each review segment is deficient, we report how well the model-generated deficiency explanation aligns with the human explanation.

\begin{table}[t]
    \centering
    \caption{Evaluation of LLMs' explanations for correctly identified deficient~segments.}
    \begin{tabular}{@{}lc@{}}
        \toprule
        {Model} & {ROUGE-1/2/L/BERTScore} \\
       \midrule
        {GPT-4} & 17.13 / 2.71 / 14.64 / 55.63 \\ 
        {Claude Opus} &  \textbf{20.18} / \textbf{3.69} / \textbf{17.52} / \textbf{57.28}  \\ 
        {Gemini 1.5} &  18.47 / 2.98 / 16.38 / 56.46  \\ 
        {Llama3-8B} & 16.49 / 2.22 / 13.65 / 55.23 \\ 
        {Llama3-70B} & 15.94 / 1.95 / 13.78 / 57.09 \\ 
        {Qwen2-72B} & 17.07 / 3.00 / 14.69 / 56.88 \\ 
        \bottomrule

    \end{tabular}
    \label{tab:meta_review_explanation}
\end{table}

We put the results of \citet{du2024llms} in Table~\ref{tab:meta_review_explanation}. The results in Table~\ref{tab:meta_review_explanation} show that overall scores for all LLMs are relatively low, indicating they can identify some Deficient segments but struggle to articulate their reasoning. Among the LLMs, Claude Opus achieves the highest scores across all metrics, suggesting its explanations align best with human annotators. Claude Opus also excels in identifying Deficient segments, as shown previously. GPT-4 and Gemini 1.5 show similar performance to Claude Opus. The open-source models, Llama3 (8B and 70B) and Qwen2-72B, generally score lower than the closed-source models.

\section{Conclusion}
 In this work, we propose \dataname, a novel benchmark targeting a comprehensive evaluation of the current LLMs' capacity in assisting AI research. \dataname~consists of distinct expertise-intensive tasks along with the curated evaluation metrics. We collect high-quality data by employing senior AI researchers and conducting strict data examinations. Extensive experiments highlight the challenges and values of \dataname.

\section*{Limitations}
We shed light on two limitations of this work:
i) As we gather data from open-source platforms such as arXiv and OpenReview, there is a possibility that current or future LLMs may be trained on the same source data utilized in our benchmark. This situation could influence the fairness of LLM comparisons and the conclusions drawn from this paper. While we acknowledge this potential data leakage, we maintain that our work can provide valuable insights and \emph{serve as an upper bound for some LLMs}, particularly if they are indeed pretrained on those scientific papers. At present, this research can inform the design of current LLMs and agents, enhancing the community's understanding of the strengths and limitations of using LLMs in scientific research.
ii) Meanwhile, due to the high consumption of employing senior researchers in conducting data annotation, the data size for some tasks, such as \tasknameexperiment, is relatively small. This might introduce variability in LLM performance, even with repeated runs. As this work marks the beginning of a series, we plan to release larger-scale datasets that will cover more diverse research tasks in AAAR-2.0.

\section*{Acknowledgments}
We would like to thank arXiv and OpenReview for releasing the paper source packages and the review comments.
The authors would also like to thank Ibraheem Moosa and Sarkar Snigdha Sarathi Das for assisting in the data collection.

\section*{Impact Statement}
Our study explores whether LLMs can assist human researchers in AI research. We do not advocate for AI replacing human researchers. Instead, we stress that the primary responsibility for scientific research should remain with humans to prevent societal risks, with LLMs serving as tools to enhance research efficiency. Specifically, our work analyzes the strengths and weaknesses of LLMs to ensure researchers remain judicious in their use of these tools. Our goal is to mitigate risks while maximizing the benefits offered by LLMs. We are committed to the careful distribution of data collected in our research, ensuring it is used solely for research purposes. 

\nocite{liu2024lost,wang2024leave}

\bibliography{main}
\bibliographystyle{icml2025}

\newpage
\appendix
\onecolumn



\section*{Appendices}
\label{sec:appendix}

Within this supplementary material, we elaborate on the following aspects:
\begin{itemize}
\item Appendix \ref{appendix:statistic}: Data Statistics and Diversity
\item Appendix \ref{appendix:details}: Implementation Details
\item Appendix \ref{appendix:more_experiments}: More Experiment Results and Details
\item Appendix \ref{appendix:cases}: Data Cases and Annotation Platform Illustration
\item Appendix \ref{appendix:prompt}: Prompt Templates 

\end{itemize}

\section{Data Statistics and Diversity}
\label{appendix:statistic}

We provide the detailed data statistics of three datasets in our benchmark, as shown in Table~\ref{tab:equations_statistics}, \ref{tab:experiment_statistics}, and \ref{tab:review_weakness_statistics}. We use the NLTK package\footnote{\url{https://www.nltk.org/}} to tokenize words and count the length. When calculating the length of equations, we use the pylatexenc tool\footnote{\url{https://github.com/phfaist/pylatexenc}} to simplify the equations first.

\begin{figure}[!h]
 \centering
    \begin{subfigure}{0.69\linewidth}
        \centering
        \includegraphics[width=0.99\linewidth]{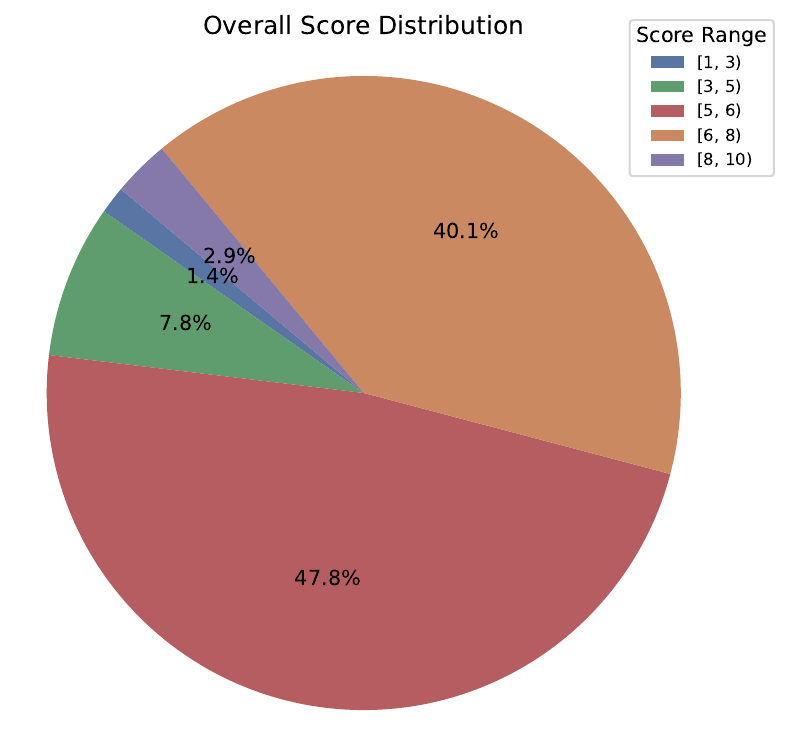}
        \caption{The review score distribution of the papers used in \protect\ColoredReview.}
        \label{fig:task3_review_score_dis}
    \end{subfigure}
    
    \vspace{6pt} 
    
    \begin{subfigure}{0.69\linewidth}
        \centering
        \includegraphics[width=0.99\linewidth]{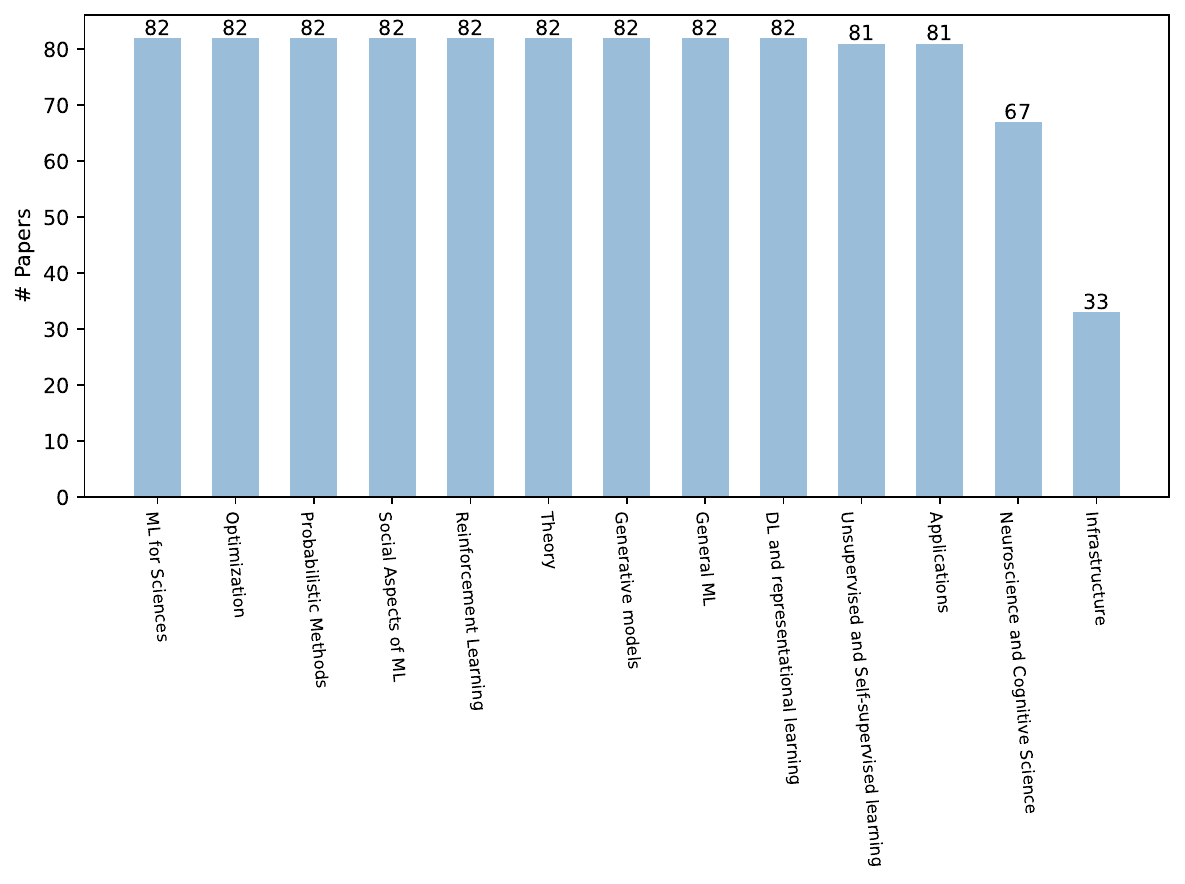}
        \caption{The track distribution of the papers used in \protect\ColoredReview.}
        \label{fig:task3_review_track_dis}
    \end{subfigure}
    
	\caption{The data diversity illustration of \protect\ColoredReview, including the score distribution and track distribution of the papers used in our dataset.}
	\label{fig:task3_data_distribution}
\end{figure}

Meanwhile, for the \tasknamereview, we also plot the review scores distribution of the papers used in the dataset, as well as the track distribution. As can be found in Figure~\ref{fig:task3_data_distribution}, our dataset has a decent distribution, where the papers are uniformly distributed across 13 tracks, and most papers' scores ranged from 5 to 8 (i.e., most papers are weakly rejected or accepted).

\begin{table}[h!]
\centering
\tiny
\caption{The statistics of \protect\ColoredEQ. Here, the ``\texttt{{left}}'' and ``\texttt{{right}}'' input context indicates the paper contexts {\ul before} and { \ul after} the missed equation; ``\texttt{pos.}'' means the ground-truth equations (written by the source paper authors), while ``\texttt{neg.}'' is the GPT4-synthetic wrong equations.}
\resizebox{0.6\textwidth}{!}{
\begin{tabular}{lr}
\toprule
\# of positive equations         & 1,049                 \\
\# of negative equations         & 3,147                 \\
\# of source papers                      & 869                   \\ \midrule
ave. ``\texttt{left}'' input context length (in words) 
   &  4,377                     \\

ave. ``\texttt{right}'' input context length (in words)
   &  6,362                     \\ 

max ``\texttt{left}'' input context length (in words)
   &  24,849                     \\

max ``\texttt{right}'' input context length (in words)
   &  32,948                    \\

min ``\texttt{left}'' input context length (in words)
   &  711                     \\

min ``\texttt{right}'' input context length (in words)
   &  8                    \\

\midrule

ave. ``\texttt{pos.}'' output equation length (in character)
   &  55                     \\

ave. ``\texttt{neg.}'' output equation length (in character)
   &  48                     \\

max ``\texttt{pos.}'' output equation length (in character)
   &  1,039                    \\

max ``\texttt{neg.}'' output equation length (in character)
   &  306                     \\

min ``\texttt{pos.}'' output equation length (in character)
   &  6                     \\

min ``\texttt{neg.}'' output equation length (in character)
   &  4                     \\

\bottomrule

\end{tabular}
}

\label{tab:equations_statistics}
\end{table}

\begin{table}[h!]
\centering
\tiny
\caption{The statistics of \protect\ColoredEXP.}
\resizebox{0.59\textwidth}{!}{
\begin{tabular}{lr}
\toprule
\# of instances         & 100                 \\
\# of source papers                      & 100                   \\ 

\midrule

ave. input context length (in words) 
   &  4,288                     \\

max input context length (in words) 
   &  9,799                     \\

min input context length (in words)
   &  698                    \\ 

ave. \# of input figures
   &  2.6                    \\

max \# of input figures
   &  16.0                    \\

min \# of input figures
   &  0.0                    \\

\midrule

ave. length of Experiment\&Explanation list
   &  5.7                     \\

ave. length per experiment (in words)
   &  34.3                     \\

ave. length per explanation (in words)
   &  27.1                    \\

max length of Experiment\&Explanation list
   &  13                    \\

max length per experiment (in words)
   &  135                     \\

max length per explanation (in words)
   &  89                     \\

min length of Experiment\&Explanation list
   &  2                     \\

min length per experiment (in words)
   &  9                     \\

min length per explanation (in words)
   &  9                     \\

\bottomrule

\end{tabular}
}
\label{tab:experiment_statistics}
\end{table}

\begin{table}[h!]
\centering
\tiny
\caption{The statistics of \protect\ColoredReview.}
\resizebox{0.53\textwidth}{!}{
\begin{tabular}{lr}
\toprule
\# of instances         & 993                 \\
\# of source papers                      & 993                   \\ 

\midrule

ave. input context length (in words) 
   &  9,811                     \\

max input context length (in words) 
   &  49,195                     \\

min input context length (in words)
   &  24                    \\ 

ave. \# of input figures
   &  7.0                    \\

max \# of input figures
   &  37.0                    \\

min \# of input figures
   &  0.0                    \\

ave. \# of input tables
   &  4.3                    \\

max \# of input tables
   &  53.0                    \\

min \# of input tables
   &  0.0                    \\

\midrule

ave. \# of reviewers per paper 
   & 3.8                      \\

max \# of reviewers per paper 
   & 9.0                      \\

min \# of reviewers per paper 
   & 3.0                      \\

ave. \# of weaknesses per reviewer
   & 4.8                       \\

max \# of weaknesses per reviewer
   & 39.0                       \\

min \# of weaknesses per reviewer
   & 1.0                       \\

ave. length of weakness (in words)
   &  39.1                     \\

max length of weakness (in words)
   &  371.0                     \\

min length of weakness (in words)
   &  2.0                     \\

\bottomrule

\end{tabular}
}

\label{tab:review_weakness_statistics}
\end{table}

\section{Implementation Details}
\label{appendix:details}

\subsection{Metric Details}
\label{appendix:details_metric}

When calculating the metrics, specifically for the similarity-based scores, we utilize SentenceBERT~\citep{reimers2019sentence} to encode each segment (e.g., each experiment idea in the list) into a dense vector, and then calculate the cosine similarity,\footnote{\url{https://huggingface.co/sentence-transformers/all-mpnet-base-v2}} which takes about 1GB of memory when running on a single A100 GPU.

\subsection{LLMs Running Details}
\label{appendix:details_model}

In our experiments, we utilize various LLMs, including both closed and open-sourced. We list the model weight sources for the open-source LLMs:
\begin{itemize}
    \item OLMo-7B~\citep{Groeneveld2023OLMo}: \url{https://huggingface.co/allenai/OLMo-7B}
    \item Falcon-40B~\citep{falcon40b}: \url{https://huggingface.co/tiiuae/falcon-40b} 
    \item Gemma 2-27B~\citetalias{gemma_2024}: \url{https://huggingface.co/google/gemma-2-27b}
    \item Mistral-7B~\citep{jiang2023mistral}: \url{https://huggingface.co/mistralai/Mistral-7B-Instruct-v0.3}
    \item Mixtral-8x22B-MoE~\citep{jiang2024mixtral} : \url{https://huggingface.co/mistralai/Mixtral-8x22B-Instruct-v0.1}
    \item Llama 3.1-70B~\citep{Llama3.1:online}: \url{https://huggingface.co/meta-llama/Llama-3.1-70B}
    \item Qwen 2.5-72B~\citetalias{qwen2.5}: \url{https://huggingface.co/Qwen/Qwen2.5-72B}
\end{itemize}

We use VLLM to unify the inference endpoints of all the above models.\footnote{\url{https://github.com/vllm-project/vllm}} We use Pytorch 2.4.0 with CUDA 12.1, and use 8 NVIDIA A100 GPUs for the LLMs inference.

Meanwhile, we use the {gpt-4o-2024-08-06}, {gpt-4-1106-preview}, {o1-preview-2024-09-12}, {gemini-1.5-pro-002}, and {claude-3-5-sonnet-20240620} for the closed-source LLMs. We use LiteLLM to unify the API calling for all these LLMs.\footnote{\url{https://github.com/BerriAI/litellm}} 

Given the unstable performance of LLMs, particularly closed-source ones, we run each model thrice during our experiments, selecting the median result from these repeated runs.


\section{More Experiment Results and Details}
\label{appendix:more_experiments}

\subsection{Input Context Scaling Investigation}
\label{appendix:more_exp_context_scaling}

\begin{figure}[!h]
 \setlength{\belowcaptionskip}{-12pt}
 \setlength{\abovecaptionskip}{0.6pt}
	\begin{center}
		\centering
		\includegraphics[width=0.48\textwidth]{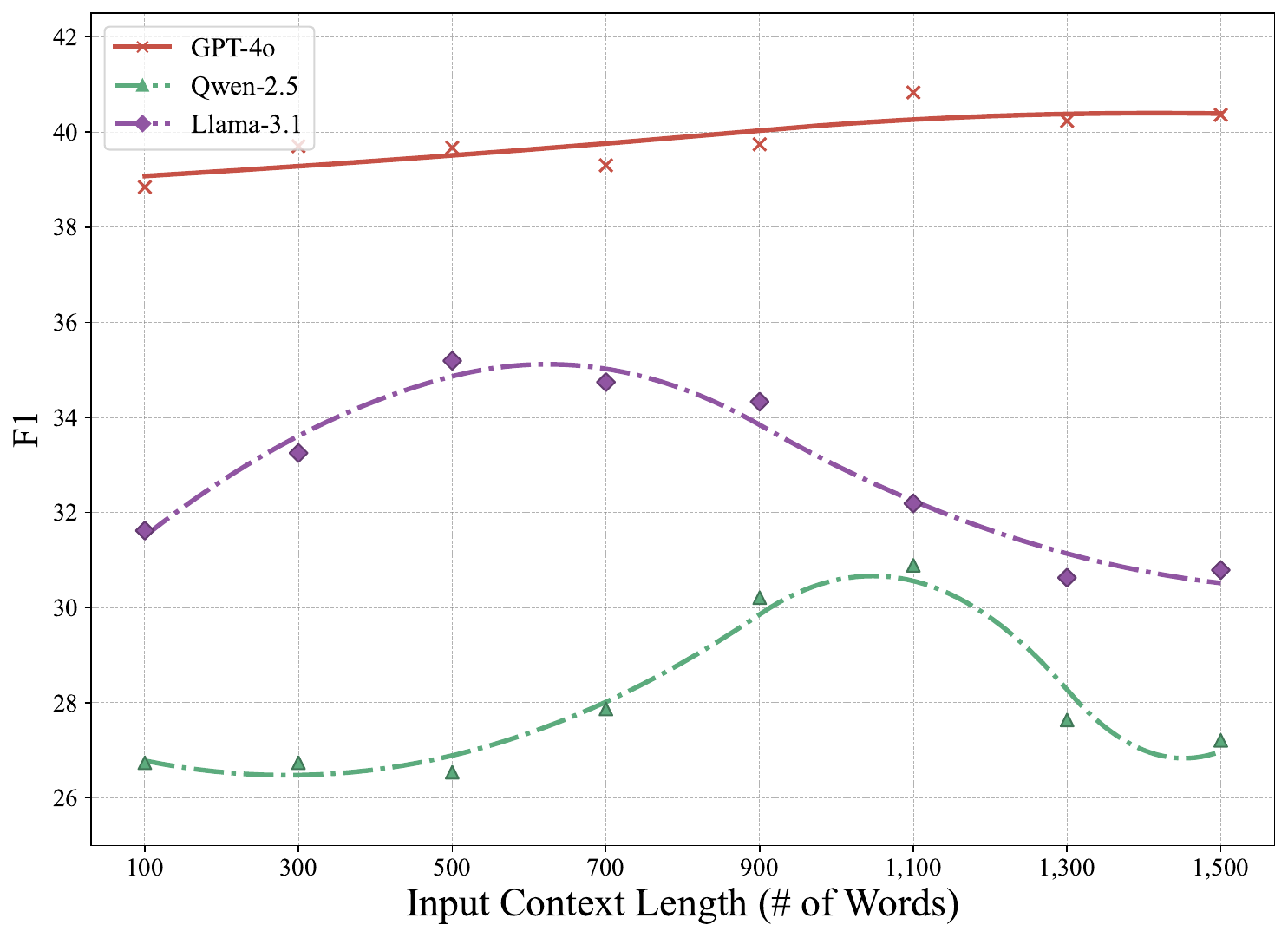}
	\end{center}
	\caption{The input context length scaling trend on the~\protect\ColoredEQ~task.
 }
\label{fig:task1_equation_scaling}
\end{figure}

\begin{figure}[!h]
 \centering
    \begin{subfigure}{0.49\linewidth}
        \centering
        \includegraphics[width=0.99\linewidth]{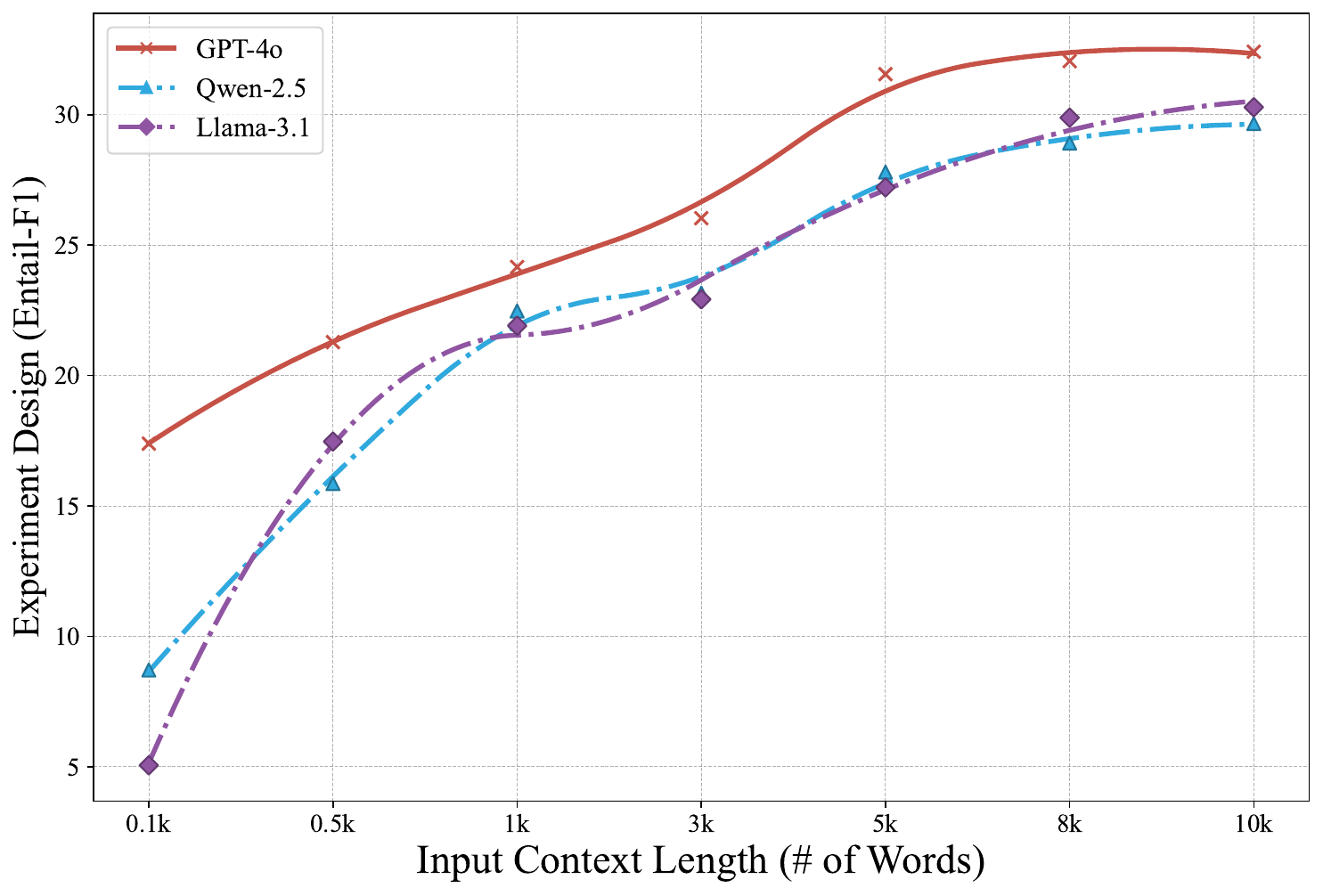}
        \label{fig:task2_f1}
    \end{subfigure}
    
    \vspace{-1pt} 
    
    \begin{subfigure}{0.49\linewidth}
        \centering
        \includegraphics[width=0.99\linewidth]{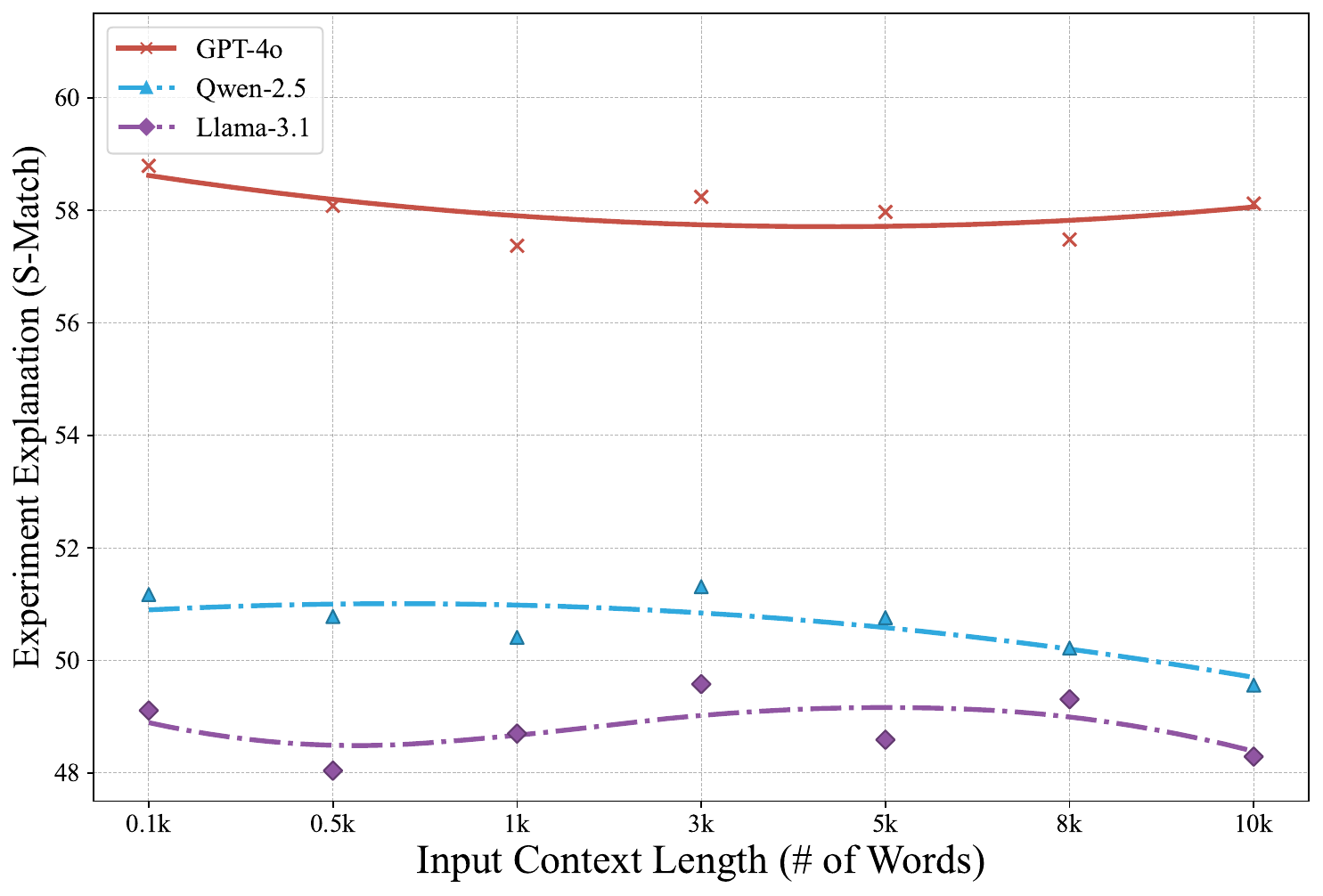}
        \label{fig:task2_match}
    \end{subfigure}
    
	\caption{The input context length scaling trend of different LLMs on the~\protect\ColoredEXP~task.}
	\label{fig:task2_context_scaling}
\end{figure}

Figure~\ref{fig:task1_equation_scaling}, Figure~\ref{fig:task2_context_scaling}, and Table~\ref{tab:review_context} show the context scaling results of \tasknameequation, \tasknameexperiment, and \tasknamereview.

\begin{table*}[th!]
\centering
\tiny
\caption{The performance comparison of different input processing methods for \protect\ColoredReview. We use GPT-4o and GPT-4-Turbo because both accept a maximum of 128k tokens input. We also put the results of \AISci~in the table for reference. Here, ``split-combine'' splits the input paper into several pieces, where each piece's length is denoted as ``window size''; ``no-split'' means the conventional input cutting, for example, if the window size is 3,000, then only the first 3,000 words in the paper are used. According to the data statistics, 20,000 words can cover maximum lengths of more than 95\% of the papers in our dataset.}
\resizebox{0.86\textwidth}{!}{

\begin{tabular}{lcrrrrr}

\toprule

\multicolumn{1}{c}{\multirow{2}{*}{\textbf{Models}}} & \multirow{2}{*}{\textbf{\begin{tabular}[c]{@{}c@{}}Input Context\\       Processing\end{tabular}}} & \multicolumn{1}{c}{\multirow{2}{*}{\textbf{\begin{tabular}[c]{@{}c@{}}Window Size\\      (in words)\end{tabular}}}} & \multicolumn{1}{c}{\multirow{2}{*}{\textbf{\MetricReviewF}}} & \multicolumn{1}{c}{\multirow{2}{*}{\textbf{\MetricReviewPrec}}} & \multicolumn{1}{c}{\multirow{2}{*}{\textbf{\MetricReviewRecall}}} & \multicolumn{1}{c}{\multirow{2}{*}{\textbf{\MetricReviewIDF}}} \\
\multicolumn{1}{c}{}                                 &                                                                                                    & \multicolumn{1}{c}{}                                                                                                & \multicolumn{1}{c}{}                                & \multicolumn{1}{c}{}                                       & \multicolumn{1}{c}{}                                    & \multicolumn{1}{c}{}                                  \\ \midrule
\multirow{3}{*}{GPT-4o}                              & split-combine                                                                                     & 3,000                                                                                                                  & \textbf{47.73}                                               & 42.09                                                      & \textbf{55.48}                                                   & 5.95                                                  \\
                                                     & no-split                                                                                           & 3,000                                                                                                                  & 45.74                                               & \textbf{43.45}                                                    & 48.54                                                   & 5.92                                                  \\
                                                     & no-split                                                                                           & 20,000                                                                                                                 & 45.47                                               & 42.97                                                      & 48.51                                                   & \textbf{6.02}                                                  \\ \midrule
\multirow{3}{*}{\AISci}                        & split-combine                                                                                     & 3,000                                                                                                                  & \textbf{45.05}                                               & 40.02                                                      & \textbf{51.91}                                                   & 2.23                                                  \\
                                                     & no-split                                                                                           & 3,000                                                                                                                  & 42.56                                               & \textbf{40.90}                                                      & 44.65                                                   & 2.53                                                  \\
                                                     & no-split                                                                                           & 20,000                                                                                                                 & 42.53                                               & 40.75                                                      & 44.78                                                   & \textbf{2.58}                                                  \\ \bottomrule
\end{tabular}

}

\label{tab:review_context}
\end{table*}

\subsection{Human Evaluation on LLM-Generated Novel Experiments}
\label{appendix:details_human_eval_experiment_necessity}

\begin{figure*}[!th]
	\begin{center}
		\centering
		\includegraphics[width=\linewidth]{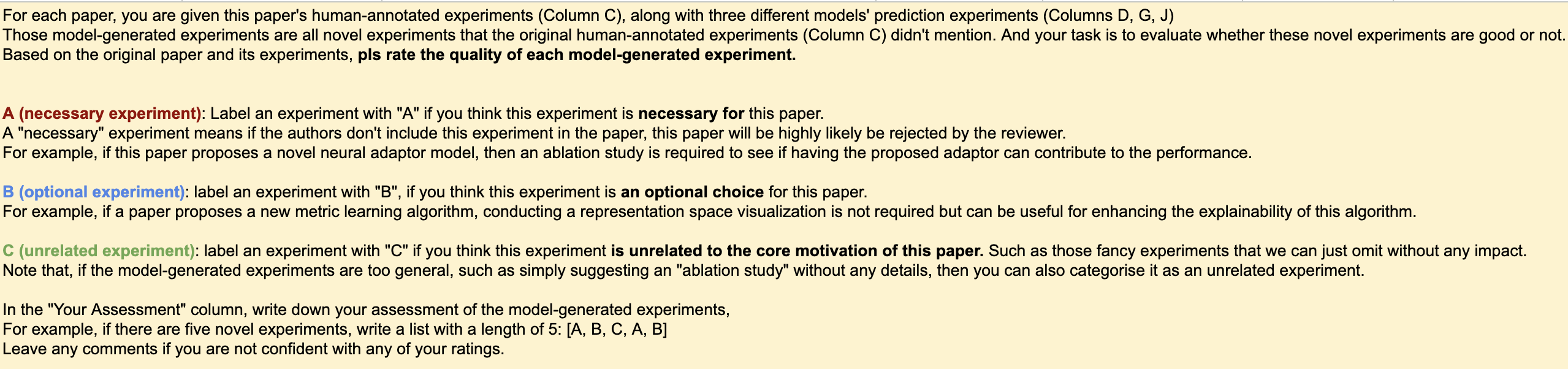}
	\end{center}
	\caption{The human guideline for evaluating the LLM-generated novel experiments.}
	\label{fig:exp_necessity_human_guidline}
\end{figure*}

Figure~\ref{fig:exp_necessity_human_guidline}~illustrates the evaluation guideline for novel experiments generated by LLMs. We ask 3 senior PhD students to evaluate each paper; that is, if the first two annotators disagree with each other, a third annotator will make a final decision. Table~\ref{tab:exp_necessity_cases} presents several human evaluation cases.

\begin{table}[th!]
\caption{Examples of human evaluation on the model-generated novel experiments.}
\label{tab:exp_necessity_cases}
\resizebox{0.98\linewidth}{!}{

\begin{tabular}{p{0.3\textwidth}|p{0.58\textwidth}|p{0.4\textwidth}|c}
\toprule
\multicolumn{1}{c|}{\textbf{Paper   Title}}                                                               & \multicolumn{1}{c|}{\textbf{Original Experiments (by human)}}                                                                                                                                                                                                                                                                                                                                                                                                                                                                                                                                                                                                                                                                                                                                                                                                                                                                                                                                                                                                                                                                                                                                                                                                                                                                                                                                                               & \multicolumn{1}{c|}{\textbf{Novel Experiment (by LLMs)}}                                                                                                                                                                    & \textbf{Rating} \\ \midrule
\begin{tabular}[c]{@{}l@{}}\parbox{0.30\textwidth}{ \textit{WiCE: Real-World   Entailment for Claims in Wikipedia}}\end{tabular}
& 

\begin{tabular}[c]{@{}l@{}}\parbox{0.50\textwidth}{1. Analysis in   Verification Problem Distribution: This paper should provide detailed   analysis and statistics about the verification problems in the proposed   dataset.\\      \\      2. Off-the-shelf entailment classification performance: The authors should   provide entailment classification performance of existing models on the   proposed dataset without fine-tuning.\\      \\      3. Human Performance: The authors should show human performance on the   proposed dataset.\\      \\      4. Performance of fine-tuned models: The authors should provide the   performance of models fine-tuned on the proposed dataset.\\      \\      5. Performance on the evidence retrieval task: The authors should show the   performance on the evidence retrieval task, which is a sub-task of the   proposed dataset.\\      \\      6. Performance of LLMs: The authors should provide the performance of LLMs   on the proposed dataset.\\      \\      7. Retrieval+Entailment: Authors should provide experiments on a framework   of retrieving evidence sentences and evaluate entailment by using the   retrieved sentences.\\      \\      8. Analysis of Claim-Split on Downstream Tasks: The authors should analyze   how claim-split, the proposed method, is effective on tasks other than the   proposed dataset.}\end{tabular}

& 

\begin{tabular}[c]{@{}l@{}}\parbox{0.40\textwidth}{ Assess model performance on WiCE without fine-tuning to test domain   generalization from traditional NLI datasets.
}\end{tabular}

& A               \\ \midrule

\begin{tabular}[c]{@{}l@{}}\parbox{0.30\textwidth}{ \textit{MetaMath:   Bootstrap Your Own Mathematical Questions for Large Language Models}}\end{tabular}
& 
\begin{tabular}[c]{@{}l@{}}\parbox{0.50\textwidth}{1. Results of multiple   LLMs on popular math datasets: The authors should show the performance of   multiple LLMs fine-tuned on their dataset on popular math datasets.\\      \\      2. Performance on open-source models with different sizes: The authors should   show the performance of models with different sizes trained on the proposed   dataset.\\      \\      3. Comparison to SOTA closed-source models: The authors show compare the   performance of open-source models trained on the proposed dataset and strong close-source   models.\\      \\      4. Evaluate the effect of augmentations: The authors need to perform an   ablation study to compare the different argumentation methods they   proposed.\\      \\      5. Analyze Training on Incorrect Answers: The authors should analyze   whether wrong answers generated in data augmentation can harm the   performance.\\      \\      6. Evaluate other ways to increase the size of training data: The authors   should evaluate other ways to increase the training data size and compare the   performance with models trained on their proposed train data.\\      \\      7. Error Analysis: The authors should analyze the performance of their   models in different conditions (e.g., lengths of questions).}\end{tabular}

                                    & \begin{tabular}[c]{@{}l@{}}\parbox{0.40\textwidth}{Prompt Sensitivity Analysis: Evaluate the sensitivity of MetaMath to different prompt formats or phrasings of mathematical questions.}\end{tabular}                                                       & 
B               \\ 

\midrule

\begin{tabular}[c]{@{}l@{}}\parbox{0.30\textwidth}{ \textit{Large Language Models Cannot Self-Correct Reasoning Yet}}\end{tabular}
& 
\begin{tabular}[c]{@{}l@{}}\parbox{0.50\textwidth}{1. Self-Correction with Oracle Labels: The authors should evaluate self-correction performance with oracle labels.

2. Intrinsic Self-Correction: The authors should show performance without using the oracle labels.

3. Analysis of Mistakes in Self-Correction: The authors should analyze the properties of mistakes made in the self-correction framework.

4. Multi-Agent Debate: The authors should evaluate self-correction with multi-agent debate.

5. Prompt Design Analysis: The authors should analyze the influence of prompt design for the initial responses on self-correction performance.}\end{tabular}

                                    & \begin{tabular}[c]{@{}l@{}}\parbox{0.40\textwidth}{Visualization of learned representations or attention mechanisms to provide insights into the model's inner workings.}\end{tabular}                                                       & 
C               \\ 
                                                                                                          \bottomrule
\end{tabular}

}
\end{table}

\subsection{Human Evaluation on LLM-Generated Explanation}
\label{appendix:details_human_eval}

We ask 5 annotators to evaluate the LLM-generated explanations. Specifically, each of them is assigned 4 or 5 papers, along with the corresponding experiment lists. For each paper, the annotator is given 5 different models' outputs (model names are anonymized), and the annotator has to decide if each LLM-generated explanation is acceptable according to the experiment.
We show the human evaluation results in Table~\ref{tab:experiment_human_eval}.

\subsection{Multi-Modal Input Ablation}
\label{appendix:more_exp_multi_modal}

We post the multi-modal ablation study of \tasknameexperiment~and~\tasknamereview~in Table~\ref{tab:experiment_multi_modal} and Table~\ref{tab:review_multi_modal}.

\begin{table*}[th!]
\centering
\tiny
\caption{The figure inputs ablation of \protect\ColoredEXP. For the maximum text input length, same as the setting in Table~\ref{tab:experiment_explanation_main_tab}, we use 2,000 and 3,000 words for open- and closed-source models, respectively. For the closed-source GPT-4o and GPT-4, as they have long context window sizes, we use all the figures of each paper. While for InternVL2, we randomly select two figures per input paper.}
\resizebox{0.83\textwidth}{!}{

\begin{tabular}{lrrrrrr}

\toprule

                                   & \multicolumn{3}{c}{ \textbf{Experiment Design}}                                                                                                                     & \multicolumn{3}{c}{ \textbf{Experiment Explanation}}                                                                                                                \\ \cmidrule(r){2-4} \cmidrule(l){5-7} 

\multirow{-2}{*}{\textbf{Models}} & \multicolumn{1}{c}{\textbf{\MetricExpF}} & \multicolumn{1}{c}{ \textbf{\MetricExpPrec}} & \multicolumn{1}{c}{ \textbf{\MetricExpRecall}} & \multicolumn{1}{c}{ \textbf{S-Match}} & \multicolumn{1}{c}{ \textbf{ROUGE-L}} & \multicolumn{1}{c}{ \textbf{ROUGE-1}} \\ \midrule
GPT-4o                                                    & 25.03	& 22.25	&\textbf{36.59}                                                         & \textbf{58.54}                                                        & \textbf{29.25}                                                        & \textbf{35.50}                                                        \\
\multicolumn{1}{r}{w/ figures}                            & \textbf{25.39}	 & \textbf{24.35}	&32.80                                                         & 58.53                                                        & 27.87                                                        & 34.30                                                        \\ \midrule


InternVL2-26B                                             & \textbf{24.26}	& \textbf{39.50}	& \textbf{14.91}                                                         & 50.03                                                        & 29.13                                                      & \textbf{34.26}                                                        \\
\multicolumn{1}{r}{w/ figures}                            & 15.04	& 38.50	& 8.64                                                         & \textbf{50.29}                                                        & \textbf{29.29}                                                        & 34.06                                                        \\ \bottomrule
\end{tabular}

}

\label{tab:experiment_multi_modal}
\end{table*}

\begin{table*}[th!]
\centering
\tiny
\caption{The ablation study about the paper tables and figures of \protect\ColoredReview. Based on the conclusion in Table~\ref{tab:review_context}, we use the ``split-combine'' to process the text input here (2,000 and 3,000 words context window size for open- and closed-source models). For GPT-4o, we use all the table/figure images; while for InternVL2, we randomly select two images per paper, i.e., two random figures, two random tables, or one random figure + table.}
\resizebox{0.7\textwidth}{!}{

\begin{tabular}{llrrrr}
\toprule
\multicolumn{2}{l}{\multirow{2}{*}{\textbf{Models}}} & \multicolumn{1}{c}{\multirow{2}{*}{\textbf{\MetricReviewF}}} & \multicolumn{1}{c}{\multirow{2}{*}{\textbf{\MetricReviewPrec}}} & \multicolumn{1}{c}{\multirow{2}{*}{\textbf{\MetricReviewRecall}}} & \multicolumn{1}{c}{\multirow{2}{*}{\textbf{\MetricReviewIDF}}} \\
\multicolumn{2}{l}{}                                 & \multicolumn{1}{c}{}                                & \multicolumn{1}{c}{}                                       & \multicolumn{1}{c}{}                                    & \multicolumn{1}{c}{}                                  \\ \midrule
\multicolumn{2}{l}{GPT-4o}                           & \textbf{47.73}                                      & \textbf{42.09}                                             & \textbf{55.48}                                          & \textbf{5.95}                                         \\
                & w/ tables                          & 46.76                                               & 41.32                                                      & 54.17                                                   & 5.53                                                  \\
                & w/ figures                         & 46.62                                               & 41.20                                                      & 54.04                                                   & 5.48                                                  \\
                & w/ tables \& figures               & 46.58                                               & 41.17                                                      & 53.98                                                   & 5.36                                                  \\ \midrule
\multicolumn{2}{l}{InternVL2-26B}                    & 41.91                                               & 41.02                                                      & 43.28                                                   & \textbf{1.48}                                         \\
                & w/ tables                          & 40.55                                               & 40.37                                                      & 42.91                                                   & 1.46                                                  \\
                & w/ figures                         & \textbf{42.88}                                      & \textbf{42.10}                                             & \textbf{43.76}                                          & 1.46                                                  \\
                & w/ tables \& figures               & 42.44                                               & 42.00                                                      & 43.31                                                   & 1.44                                                  \\ \bottomrule
\end{tabular}

}

\label{tab:review_multi_modal}
\end{table*}

\section{Data cases and Annotation Platform Illustration}
\label{appendix:cases}

\begin{figure*}[!h]
	\begin{center}
		\centering
		\includegraphics[width=\linewidth]{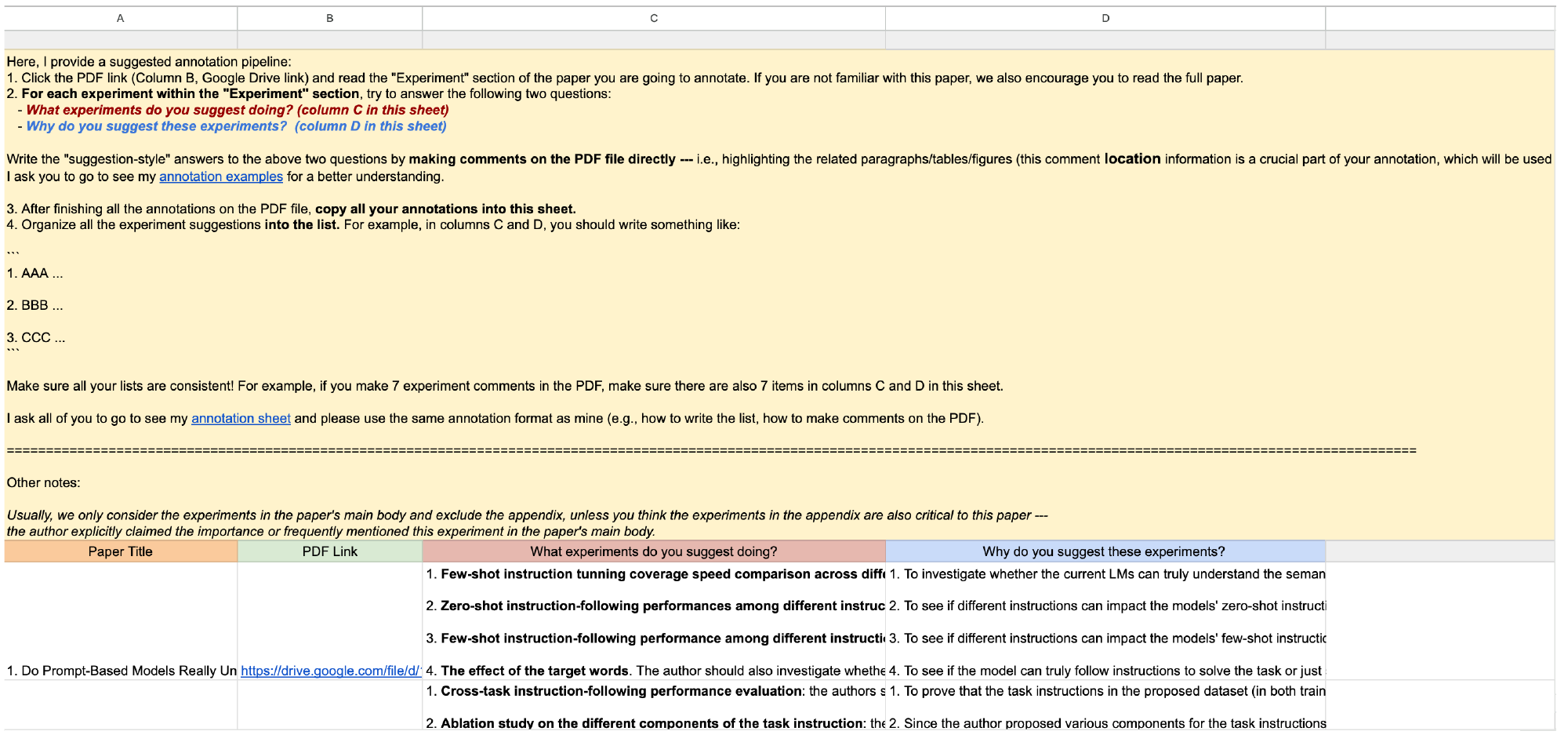}
	\end{center}
	\caption{The annotation platform for collecting the annotation of \protect\ColoredEXP. We ask annotators to first make comments on the Google Drive PDF, then move all the annotations to the online Google Doc (for further verification and discussion).}
	\label{fig:annotation_plat}
\end{figure*}

\begin{figure*}[!th]
	\begin{center}
		\centering
		\includegraphics[width=\linewidth]{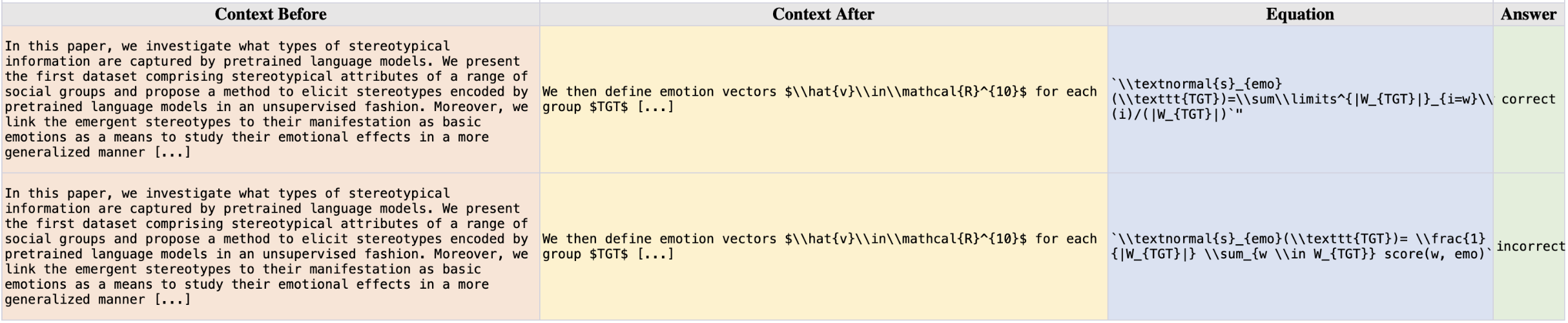}
	\end{center}
	\caption{Two sample cases of \protect\ColoredEQ.}
	\label{fig:case_eq}
\end{figure*}

\begin{figure*}[!th]
	\begin{center}
		\centering
		\includegraphics[width=\linewidth]{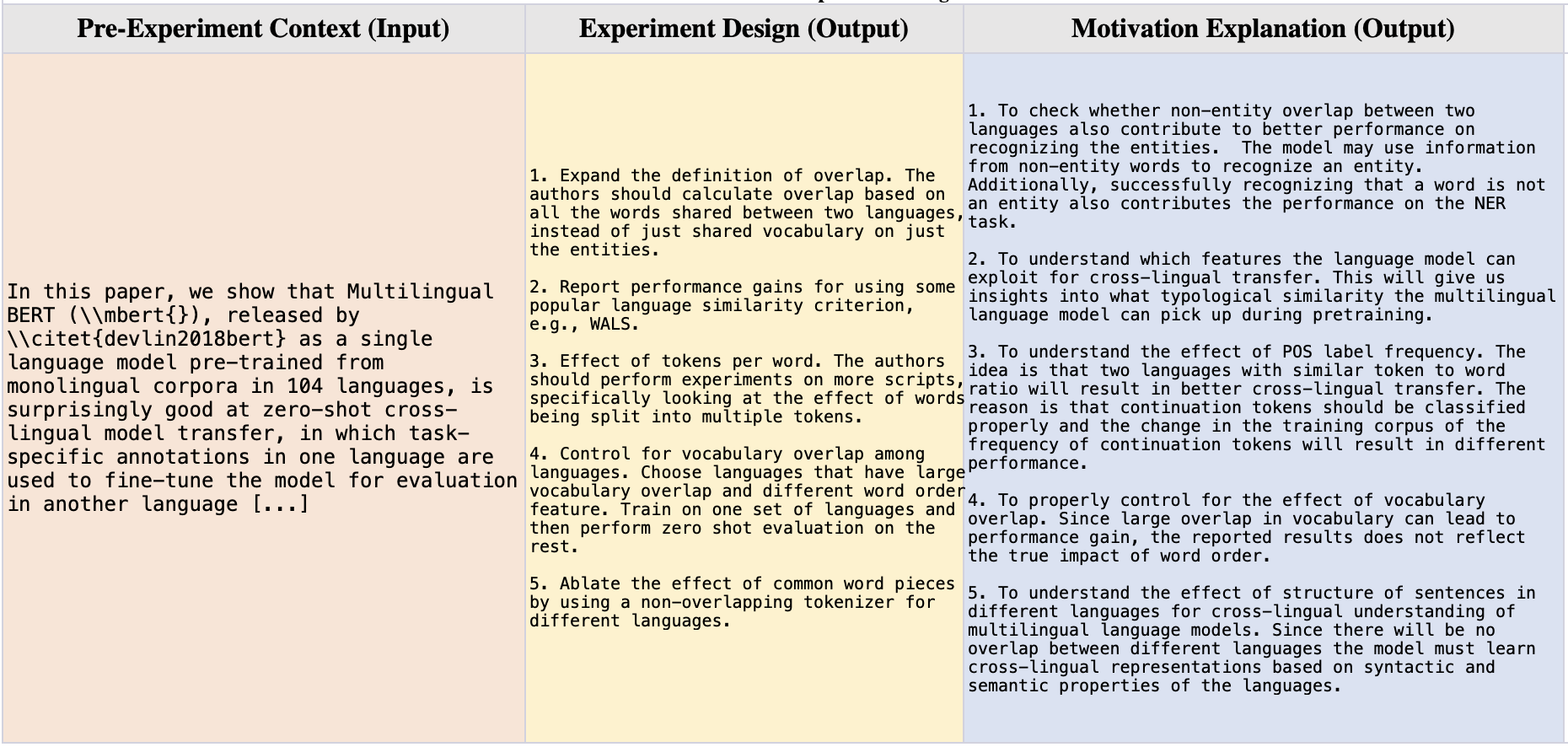}
	\end{center}
	\caption{A sample case of \protect\ColoredEXP.}
	\label{fig:case_exp}
\end{figure*}

\begin{figure*}[!th]
	\begin{center}
		\centering
		\includegraphics[width=\linewidth]{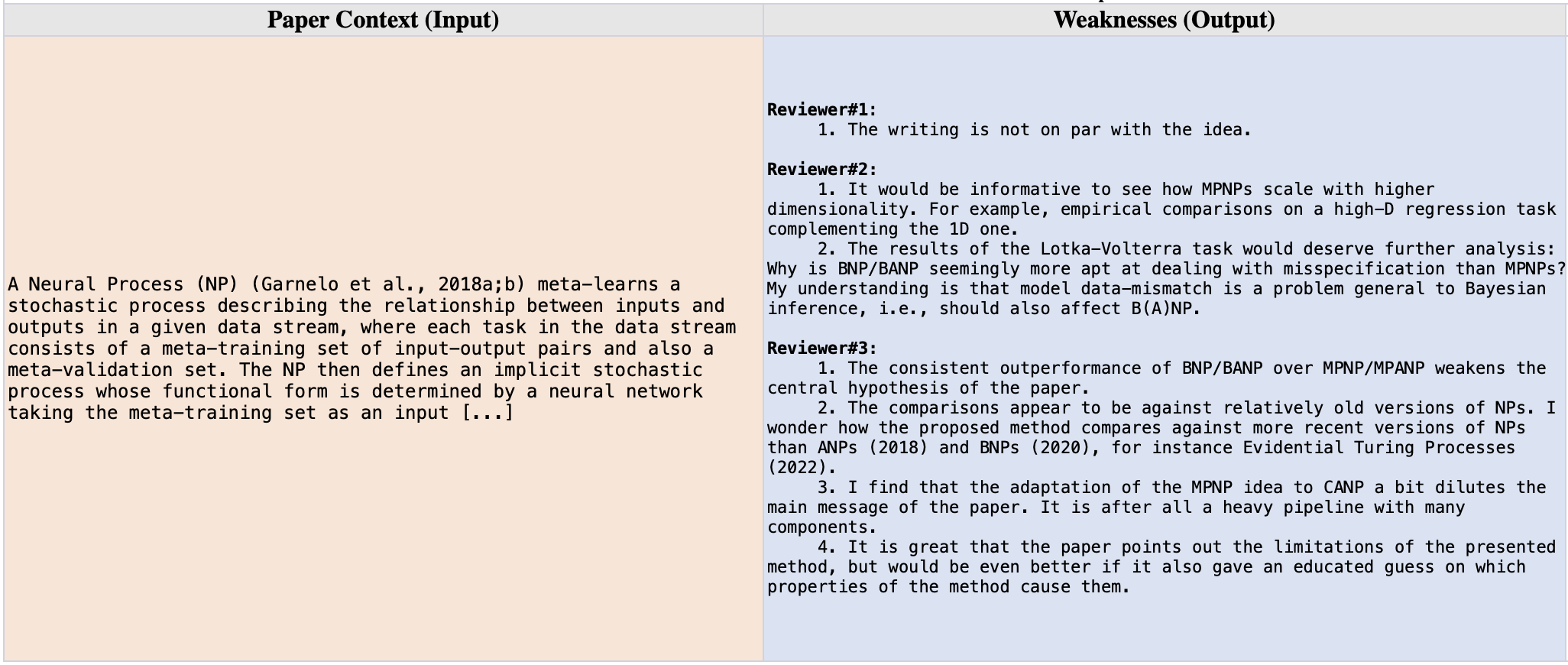}
	\end{center}
	\caption{A sample case of \protect\ColoredReview.}
	\label{fig:case_weakness}
\end{figure*}

As shown in Figure~\ref{fig:case_eq}, \ref{fig:case_exp}, and \ref{fig:case_weakness}, we show the sample cases of the three tasks in \dataname. Meanwhile, we illustrate the screenshot of our annotation platform in Figure~\ref{fig:annotation_plat}.

\section{Prompt Templates}
\label{appendix:prompt}

\begin{figure*}[!th]
	\begin{center}
		\centering
		\includegraphics[width=\linewidth]{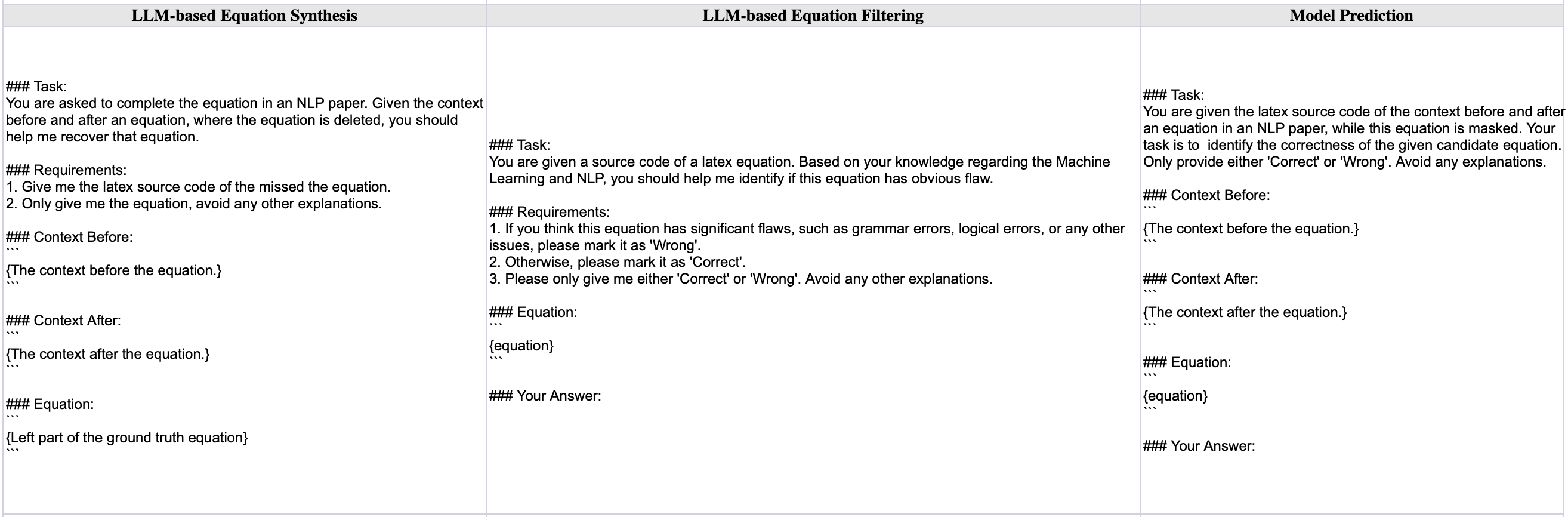}
	\end{center}
	\caption{The prompts used in \protect\ColoredEQ, including both data collection and model prediction.}
	\label{fig:prompt_equation}
\end{figure*}
\begin{figure*}[!t]
	\begin{center}
		\centering
		\includegraphics[width=\linewidth]{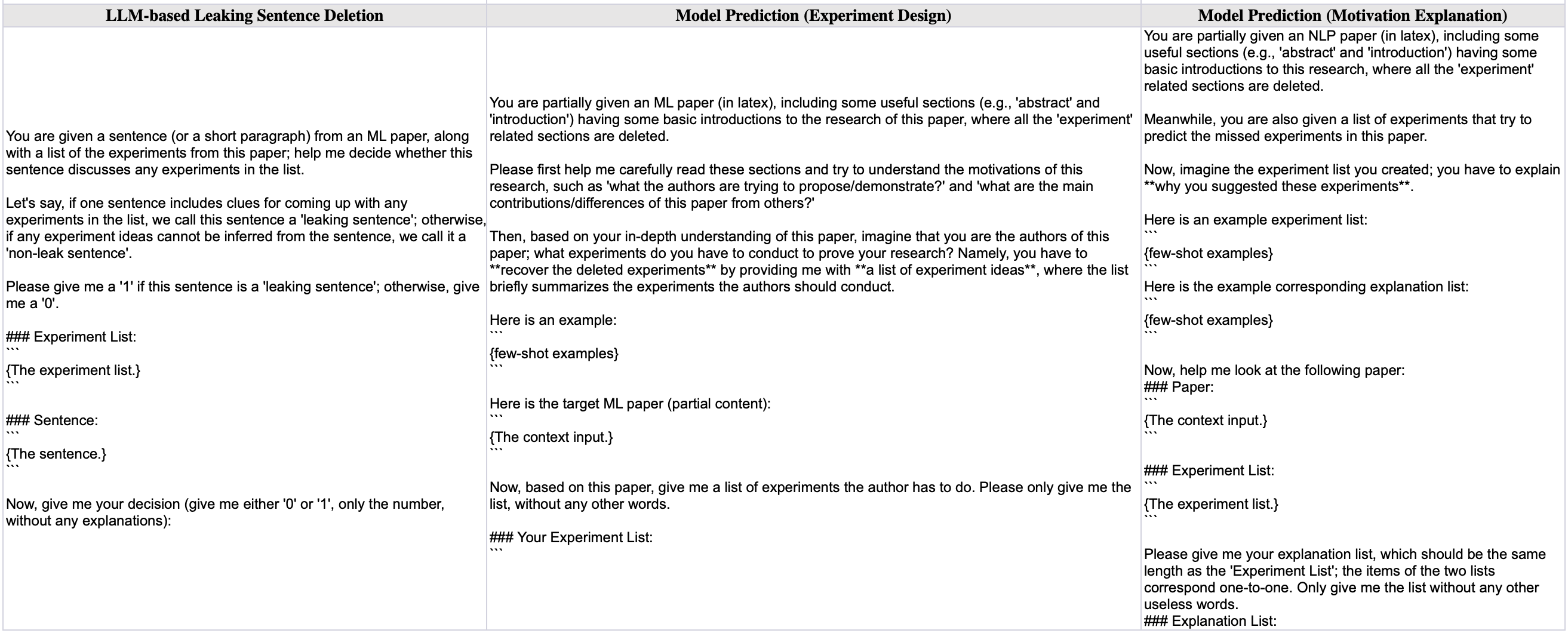}
	\end{center}
	\caption{The prompts used in \protect\ColoredEXP, including both data collection and model prediction.}
	\label{fig:prompt_exp}
\end{figure*}
\begin{figure*}[!t]
	\begin{center}
		\centering
		\includegraphics[width=0.63\linewidth]{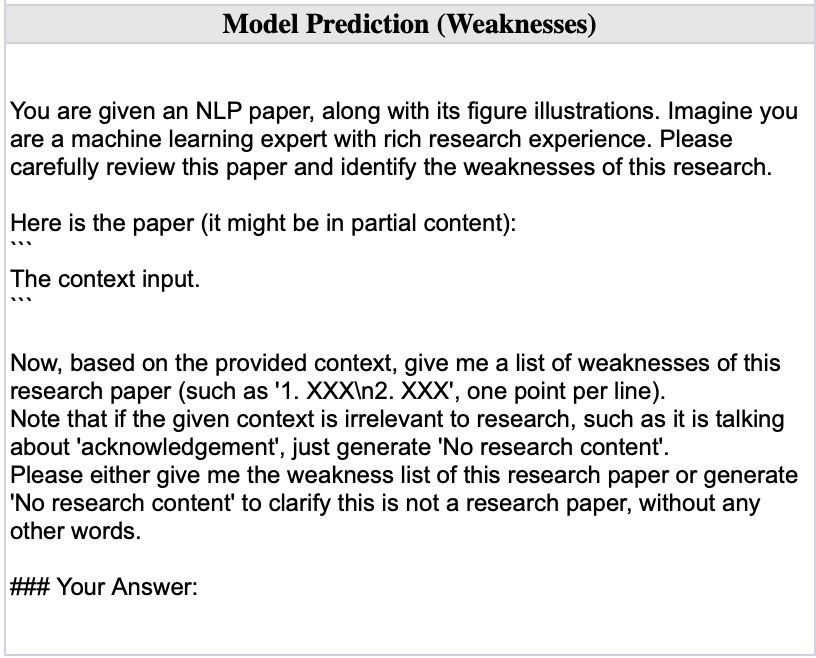}
	\end{center}
	\caption{The prompts used in \protect\ColoredReview.}
	\label{fig:prompt_weakness}
\end{figure*}

In this appendix, we attach all the prompts used in this work, including prompts in data collection and model prediction, as shown in Figure~\ref{fig:prompt_equation}, \ref{fig:prompt_exp}, and \ref{fig:prompt_weakness}.

\end{document}